\documentclass{article}
\PassOptionsToPackage{numbers, compress, sort}{natbib}
\usepackage{url}
\usepackage{graphicx}
\usepackage{multirow}
\usepackage{adjustbox}
\usepackage{wrapfig}
\usepackage{lipsum}
\usepackage{tabularx}
\usepackage{array}
\usepackage{rotating}
\usepackage{pdflscape}
\usepackage{afterpage}
\usepackage{enumitem}
\usepackage{amsmath}
\usepackage{amsthm}
\usepackage{amsfonts}
\usepackage{dsfont}
\usepackage{color}
\usepackage{subcaption}
\usepackage{booktabs}
\usepackage{hyperref}
\hypersetup{
  colorlinks   = true, 
  urlcolor     = blue, 
  linkcolor    = black, 
  citecolor   = green 
}
\usepackage[final]{flfm_neurips_2023}
\theoremstyle{plain}

\theoremstyle{definition}

\title{FedFN: Feature Normalization for Alleviating Data Heterogeneity Problem in Federated Learning}

\author{
  Seongyoon Kim\\
  Dept. ISysE, KAIST\\
  \texttt{curisam@kaist.ac.kr}\\
  \And
  Gihun Lee\\
  Graduate School of AI, KAIST\\
  \texttt{opcrisis@kaist.ac.kr}\\
  \And
  Jaehoon Oh\thanks{corresponding authors}\\
  SAIT\\
  \texttt{jh0104.oh@gmail.com}\\
  \And
  {Se-Young Yun}\footnotemark[1]\\
  Graduate School of AI, KAIST\\
  \texttt{yunseyoung@gmail.com}\\
}

\author{
  Seongyoon Kim\\
  Dept. ISysE, KAIST\\
  \texttt{curisam@kaist.ac.kr}\\
  \And
  Gihun Lee\\
  Graduate School of AI, KAIST\\
  \texttt{opcrisis@kaist.ac.kr}\\
  \And
  Jaehoon Oh\thanks{corresponding authors}\\
  SAIT
\\
  \texttt{jh0104.oh@gmail.com}\\
  \And
  {Se-Young Yun}\footnotemark[1]\\
  Graduate School of AI, KAIST\\
  \texorpdfstring{\texttt{yunseyoung@gmail.com}}{yunseyoung@gmail.com}\\
}

\begin{document}
\maketitle
\vspace{-0.24 in}
\begin{abstract}
\vspace{-0.1 in}
Federated Learning (FL) is a collaborative method for training models while preserving data privacy in decentralized settings. However, FL encounters challenges related to data heterogeneity, which can result in performance degradation. In our study, we observe that as data heterogeneity increases, feature representation in the FedAVG model deteriorates more significantly compared to classifier weight. 
Additionally, we observe that as data heterogeneity increases, the gap between higher feature norms for observed classes, obtained from local models, and feature norms of unobserved classes widens, in contrast to the behavior of classifier weight norms. This widening gap extends to encompass the feature norm disparities between local and the global models.
To address these issues, we introduce Federated Averaging with Feature Normalization Update (FedFN), a straightforward learning method. We demonstrate the superior performance of FedFN through extensive experiments, even when applied to pretrained ResNet18. Subsequently, we confirm the applicability of FedFN to foundation models. 


\end{abstract}

\vspace{-0.15 in}
\vspace{-0.05 in}
\section{Introduction}
\label{sec:intro}
\vspace{-0.1 in}
Federated Learning (FL) facilitates collaborative model training while preserving data privacy~\citep{li2020federated, mcmahan2017communication}. This approach consists of four iterative stages: (1) client selection, (2) broadcasting, (3) local training, and (4) aggregation. Selected clients receive the global model, train it locally with their own data, and transmit the trained models back to the central server for aggregation. These steps are repeated to progressively enhance the performance of global model.
In FL, a fundamental challenge arises from the existence of diverse data distributions among different clients, referred to as \emph{data heterogeneity}. This leads to performance degradation of the global model~\citep{li2021fedrs, mcmahan2017communication, li2019convergence}.


Recently, numerous studies~\citep{dong2022spherefed, oh2021fedbabu, shi2022towards} have been conducted to identify the specific aspects, such as feature representations or classifier weights, that are significantly influenced by data heterogeneity. \citet{luo2021no} demonstrated that classifier weights are the most sensitive to data heterogeneity, illustrating how classifiers can easily become biased depending on the data distribution. To mitigate classifier bias, algorithms have been proposed using restricted softmax loss or fixed orthogonal classifiers during local training~\citep{li2021fedrs, oh2021fedbabu, dong2022spherefed}. 
Meanwhile, \citet{shi2022towards} demonstrates the impact of data heterogeneity on feature representations, potentially leading to dimensional collapse where only a subset of high-dimensional feature vectors is employed to represent features within the global model. In our study, we find that the primary concern lies not in the classifier weight but in the features. 

Feature normalization, as utilized in various fields~\citep{wang2018additive, savvides2021convex, hasnat2017deepvisage, mettes2019hyperspherical, khosla2020supervised, li2021model, dong2022spherefed}, enhances the discriminative power of feature representation, making it easier to distinguish data belonging to different classes.
%
We reveal that data heterogeneity in FL leads to a substantial discrepancy in feature norms between the global model and local models. Based on this observation, we incorporate feature normalization into the FL framework. Our contributions are outlined as follows:
\begin{itemize}
    \item In FedAVG, we find that feature representations are more adversely affected by data heterogeneity than classifier weights. Furthermore, as data heterogeneity increases, the disparity between the higher feature norms for observed classes, derived from local models, and the feature norms of unobserved classes widens, in contrast to classifier weight norms. This widening gap extends to encompass feature norm disparities between local models and the global model. \textbf{(Section~\ref{sec:problem})}
    \item To tackle this challenge, we introduce \textbf{Fed}erated Averaging with \textbf{F}eature \textbf{N}ormalization Update (FedFN), which effectively eliminates discrepancies in feature norms during local training. FedFN robustly maintains the quality of feature representations even in highly heterogeneous data settings. \textbf{(Section~\ref{sec:method})}
    \item We incorporate the feature normalization technique into existing algorithms, and show notable performance improvements. Furthermore, this effectiveness persists even when using pretrained model.  \textbf{(Section~\ref{sec:exp})}
\end{itemize}



\vspace{-0.2 in}
\section{Experimental Setup}
\vspace{-0.1 in}
\noindent\textbf{Datasets and Models}
We conduct extensive experiments using two widely-used datasets with suitable models: VGG11~\citep{simonyan2014very} and ResNet18~\citep{he2016identity} for the CIFAR-10 dataset, and MobileNet~\citep{howard2017mobilenets} for the CIFAR-100 dataset~\citep{krizhevsky2009cifar}.

\noindent\textbf{Federated Environments}
To simulate a realistic FL scenario, we set the number of clients ($N$) to 100 and a fraction ratio ($r$) of 0.1 for each round of communication. Note that our investigation primarily addresses a balanced environment, wherein all clients possess datasets of identical size.\footnote{Appendix~\ref{app:add_exp} reports the results under unbalanced and non-IID derived Dirichlet distributions.}
To create data heterogeneity, we adopt a sharding partition strategy, as used in prior studies~\citep{mcmahan2017communication, oh2021fedbabu}.This strategy divides the dataset $D$ into $N \times s$ non-overlapping shards. Each client is allocated $s$ shards, with each shard containing $\frac{|D|}{N \times s}$ samples from a single class. Consequently, each client can hold samples from up to $s$ distinct classes, implying that a smaller value of $s$ leads to greater data heterogeneity. 

\noindent\textbf{Implementation Details}
Details regarding the code implementation can be found in the Appendix~\ref{app:prelim}. All experiments involve 320 communication rounds. To optimize the initial learning rate ($\eta$) and the number of local epochs ($E$) on CIFAR-10 and CIFAR-100, we conduct grid searches, with results in Appendix~\ref{app:grid search}. $\eta$ is explored in the range of $\{0.01, 0.03, 0.05, 0.1\}$ for CIFAR-10 and $\{0.1, 0.3, 0.5, 1.0\}$ for CIFAR-100. We evaluate $E$ in the set $\{1, 5, 10, 15, 20\}$ for both datasets and find optimal values of 15 for CIFAR-10 and 5 for CIFAR-100. When specific values are not mentioned, default initial learning rates of 0.01 for CIFAR-10 and 0.1 for CIFAR-100 are used.

\vspace{-0.15 in}
\section{Heterogeneity in FedAVG: The Devil is in Feature Norm Discrepancy}\label{sec:problem}
\vspace{-0.1 in}
\subsection{4-Factor Analysis of FedAVG}\label{subsec:4-factor-avg}
\vspace{-0.05 in}

Within the FedAVG, we explore the impact of data heterogeneity on both feature representations at the penultimate layer and classifier weights, denoted as $f(\cdot; \theta_{ext})\in \mathds{R}^{d}$ and $\theta_{cls}\in \mathds{R}^{C \times d}$, respectively. Our investigation centers on four factors often used to assess model performance~\citep{kang2019decoupling, papyan2020prevalence}:
\vspace{-0.25 in}
\begin{itemize}
    \item \textbf{(Factor 1) Weight similarity} quantifies the cosine similarity among classes in $\theta_{cls}$ (i.e., rows of $\theta_{cls}$). Lower values are preferred in this context.
    \item \textbf{(Factor 2) Inter-class similarity} calculates the cosine similarity among feature prototypes. A feature prototype for a class $c$ is defined as $\frac{1}{|D_{test}(c)|} \sum_{(x,y)\in D_{test}(c)}f(x; \theta_{ext})$, with a preference for lower values. Here, $D_{test}(c)$ represents the test dataset containing only samples from class $c$, and ($x,y$) denotes (input image, true class label of $x$).
    \item \textbf{(Factor 3) Intra-class similarity} quantifies the averaged cosine similarity between feature prototype and features for each class. Higher values are preferred here.
    \item \textbf{(Factor 4) Prototype-weight alignment} measures the cosine similarity between feature prototype and classifier weight for each class. Higher values are also preferred in this case.
\end{itemize}
\vspace{-0.1 in}
\newpage
These four factors encompass both feature and classifier-related aspects, enabling us to discern which aspects are more negatively impacted as data heterogeneity increases.
\begin{figure}[htp]
    \centering
    \makebox[\textwidth]{\includegraphics[width=\textwidth]{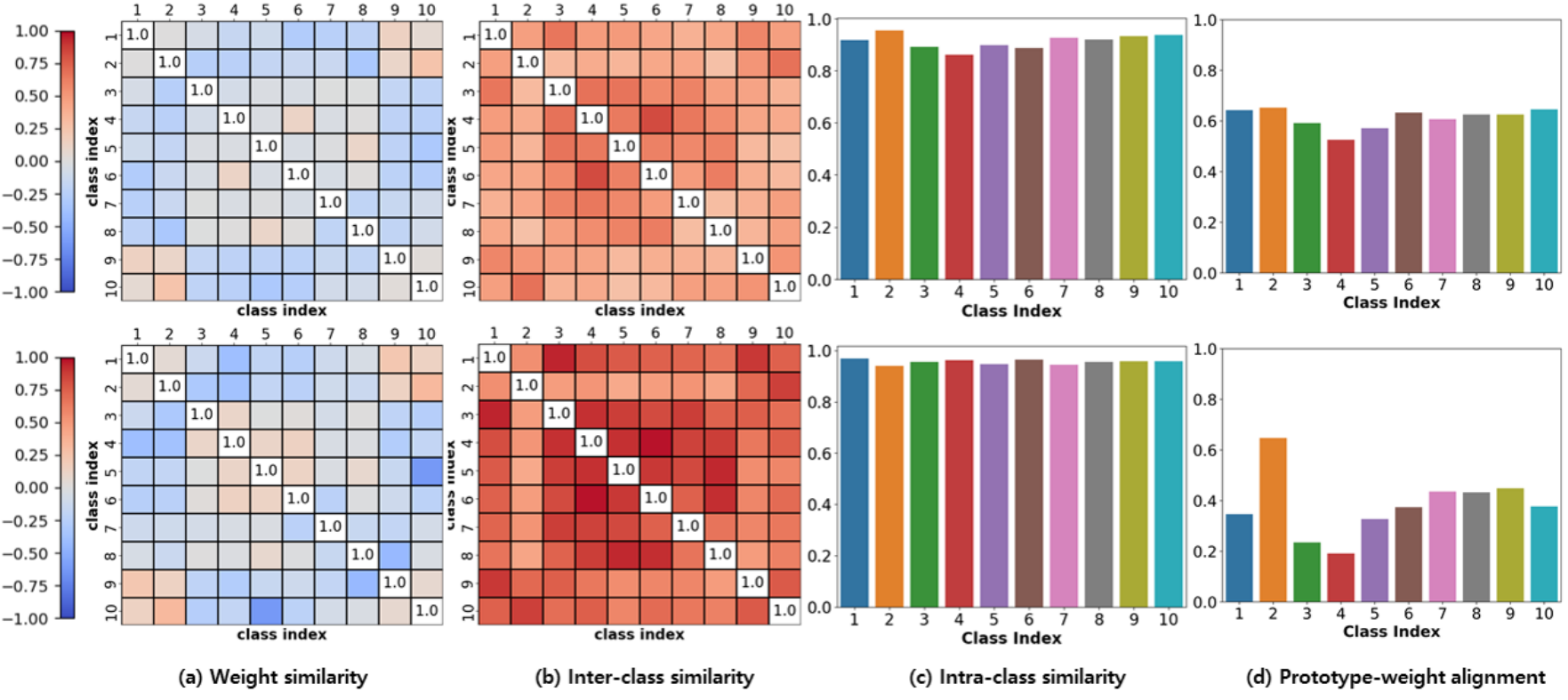}}
    \caption{4-Factor Analysis of FedAVG: Results for $s=10$ (Top Row) and $s=2$ (Bottom Row).}
    \label{fig:4_factor_fedavg}
\end{figure}
\vspace{-0.1 in}

Figure~\ref{fig:4_factor_fedavg} visualizes the four factors concerning data heterogeneity. The upper and lower rows present the results under smaller (i.e., $s$=10) and larger data heterogeneity (i.e., $s$=2), respectively. In both settings, weight similarity exhibits lower values, as indicated by the blue color. However, with increasing data heterogeneity, there is an increase in inter-class similarity within FedAVG, indicating a negative impact.  Conversely, as data heterogeneity rises, intra-class similarity improves. With higher data heterogeneity,  prototype-weight alignment deteriorates, likely influenced by the more pronounced decrease in inter-class similarity. In summary, as data heterogeneity increases, the factors most adversely affected are inter-class similarity and prototype-weight alignment, both of which are common feature-related factors.

\vspace{-0.15 in}
\subsection{Feature Norm Discrepancy Persists in Local and Global Models}
\vspace{-0.2 in}
\begin{figure}[h!]
    \centering
    \begin{subfigure}[b]{0.45\textwidth}
        \includegraphics[width=\textwidth]{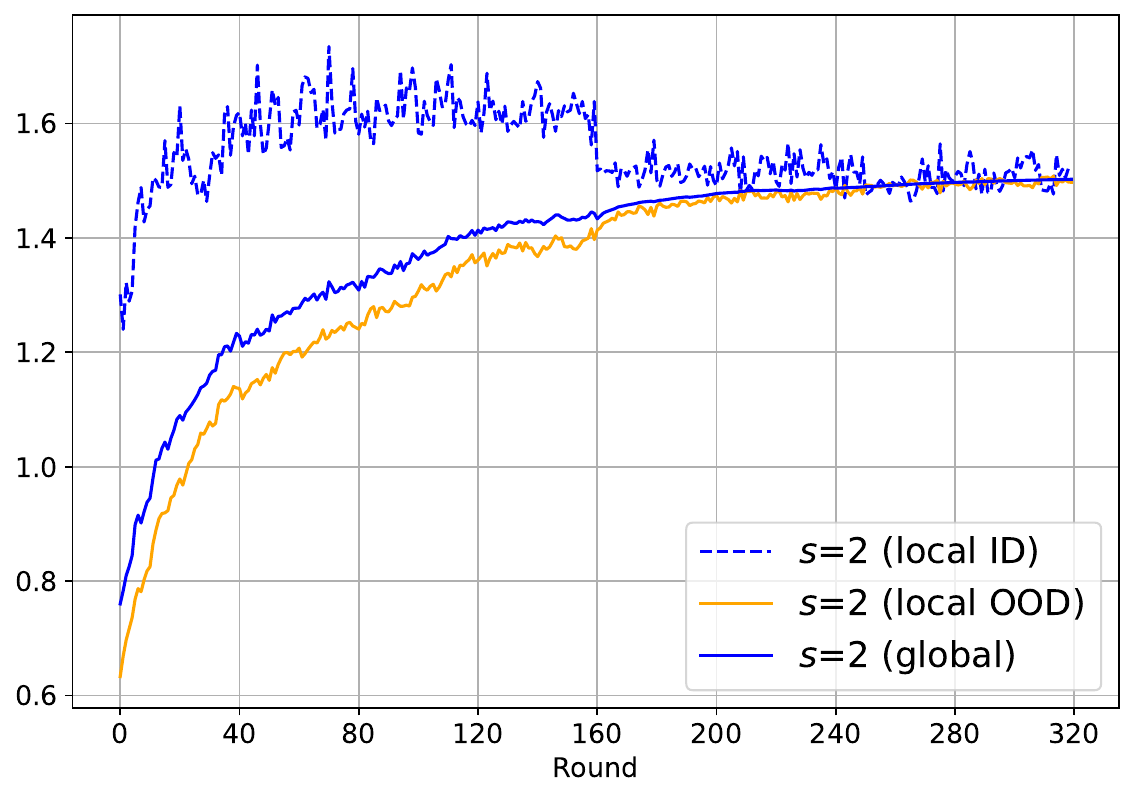}
        \label{fig:weight-norm}
        \vspace{-0.2 in}
        \caption{Classifier weight norm means}
    \end{subfigure}    
    \begin{subfigure}[b]{0.45\textwidth}
        \includegraphics[width=\textwidth]{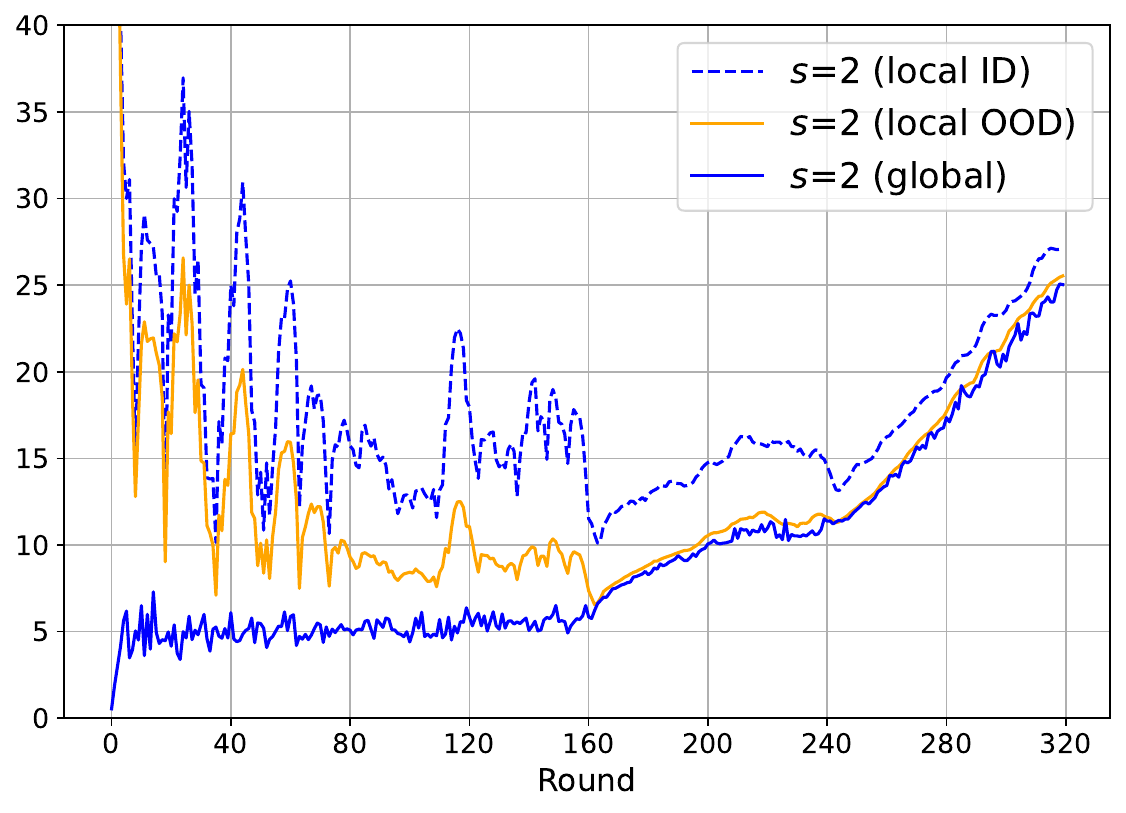}
        \label{fig:feature-norm}
        \vspace{-0.2 in}
        \caption{Feature norm means}
    \end{subfigure}
    \caption{Means of Classifier Weight and Feature Norms for Local and Global Models with $s$=2.}
    \label{fig:weight-feature-norm}
\end{figure}
\vspace{-0.1 in}
We take a closer look at feature norm bias alongside weight norm bias~\citep{luo2021no, zhao2020maintaining, wu2019large, kang2019decoupling, oh2021fedbabu, yu2020devil}, which tends to favor major classes with larger weight norms. Our investigation is conducted in a high data heterogeneity setting ($s$=2) using the CIFAR-10 dataset on VGG11. We are motivated by the extensive distribution of calssifier weight norms within classes observed across clients in FL, as discussed in \citep{luo2021no}. Furthermore, our prior four factor analysis has emphasized the significant influence of feature-related aspects in response to varying data heterogeneity. This analysis strongly encourages us to look into feature norm bias. This is important because even though a lot of research has been done on weight norm bias, not much attention has been given to feature norm bias.

We compute weight norm means for two groups of classes: those seen (ID) and unseen (OOD) classes during their respective local training. Additionally, we calculate the total class weight norm mean from the global model. Furthermore, we explore feature norm means derived from local models using both the ID and OOD test datasets, in addition to the feature norm mean obtained from the global model using the entire test dataset. Figure~\ref{fig:weight-feature-norm} illustrates the visual representation of the results.

During the initial stage of training, weight norms in local models exhibit a significant bias in favor of ID classes over OOD classes. Simultaneously, feature norm mean within local model also display a corresponding bias. However, as the learning rate gradually decreases, both feature norm bias and weight norm bias diminish. Weight norm bias eventually vanishes in local models, aligning with the weight norm mean of global model. In contrast, feature norm bias persists, consistently resulting in higher feature norm mean in local model from ID classes compared to the global model.
\vspace{-0.1 in}
\begin{wrapfigure}[13]{r}{0.5\textwidth}
    \centering
    \vspace{-0.05 in}  \includegraphics[width=\linewidth]{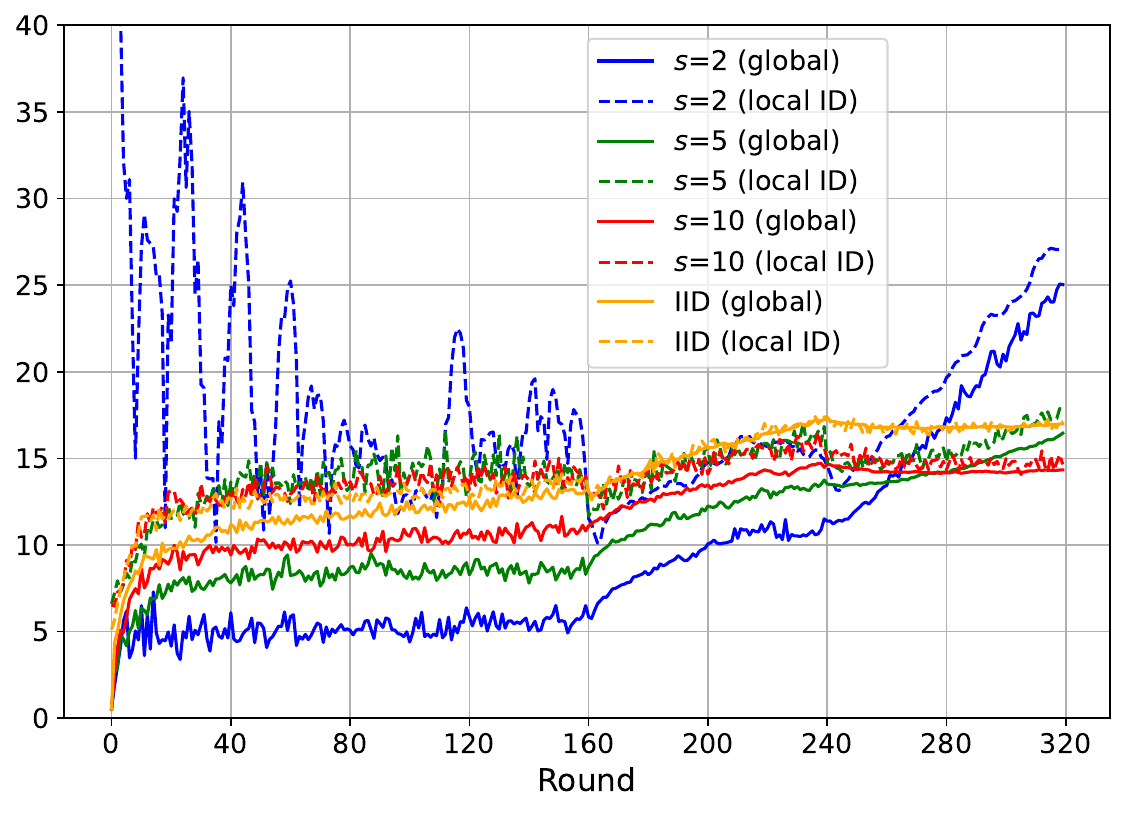}
    \vspace{-0.3 in}
    \caption{Feature Norm Means of Global and Local Models across Varying Data Heterogeneity.}
\label{fig:local_global_feature_norm}
\end{wrapfigure}
\vspace{-0.1 in}
\subsection{Feature Norm Discrepancy with Increasing Heterogeneity}
We examine the discrepancy of  feature norm means between local models on ID test dataset and global model on the entire test dataset under varying data heterogeneity. Specifically, we consider $s \in \{2, 5, 10\}$ and IID (Exactly class balanced data distribution across clients) on the CIFAR-10 dataset. As depicted in Figure~\ref{fig:local_global_feature_norm}, the discrepancy in the norms between the global model and local models increases, as data heterogeneity increases (i.e., IID $\rightarrow$ $s$=10 $\rightarrow$ $s$=5 $\rightarrow$ $s$=2). 
\vspace{-0.05 in}
\subsection{Reducing Feature Norm Discrepancy for Improved Performance}
\vspace{-0.2 in}
\begin{figure}[h!]
    \centering
    \begin{subfigure}[b]{0.48\textwidth}
        \includegraphics[width=\textwidth]{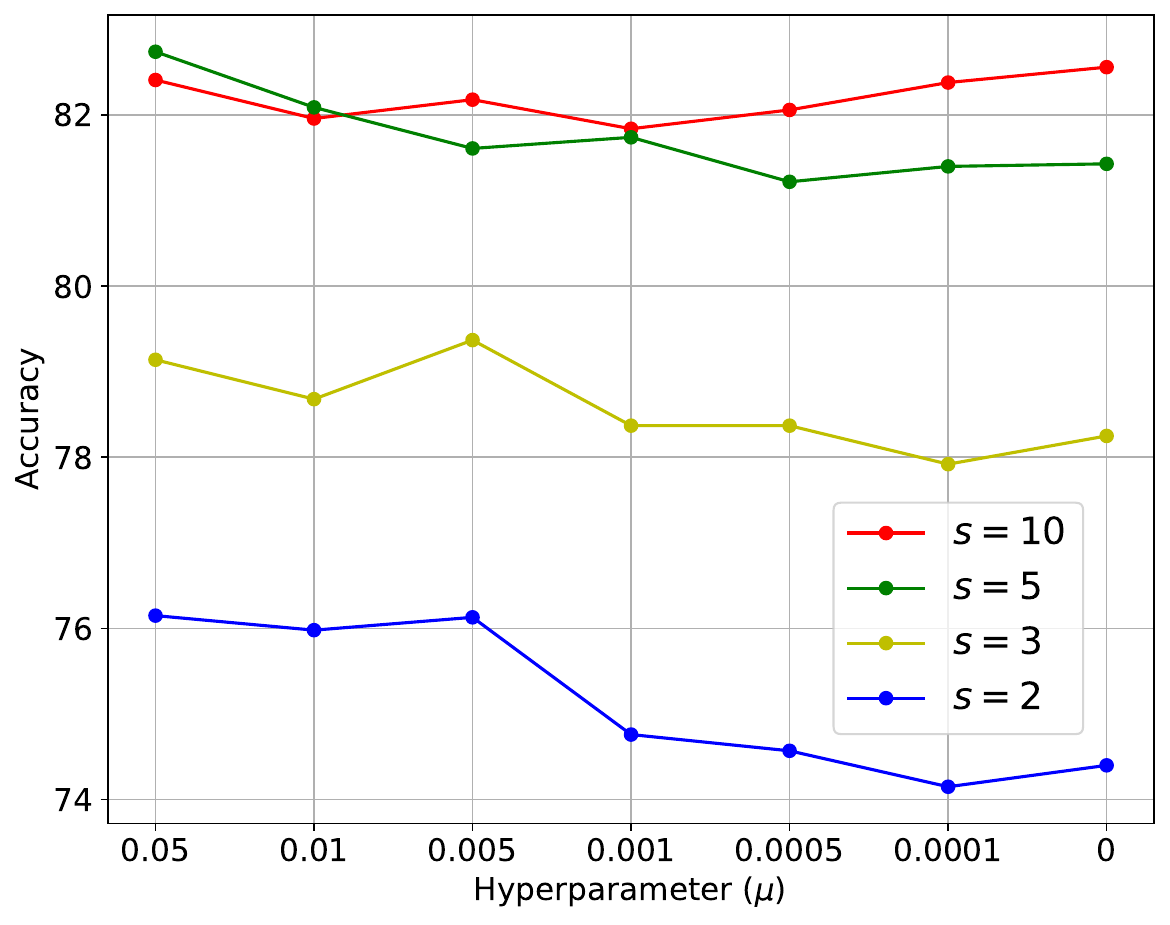}
        \label{fig:hyperparam-effect}
        \vspace{-0.25 in}
        \caption{Accuracy on hyperparameter $\mu$}
    \end{subfigure}    
    \begin{subfigure}[b]{0.47\textwidth}
        \includegraphics[width=\textwidth]{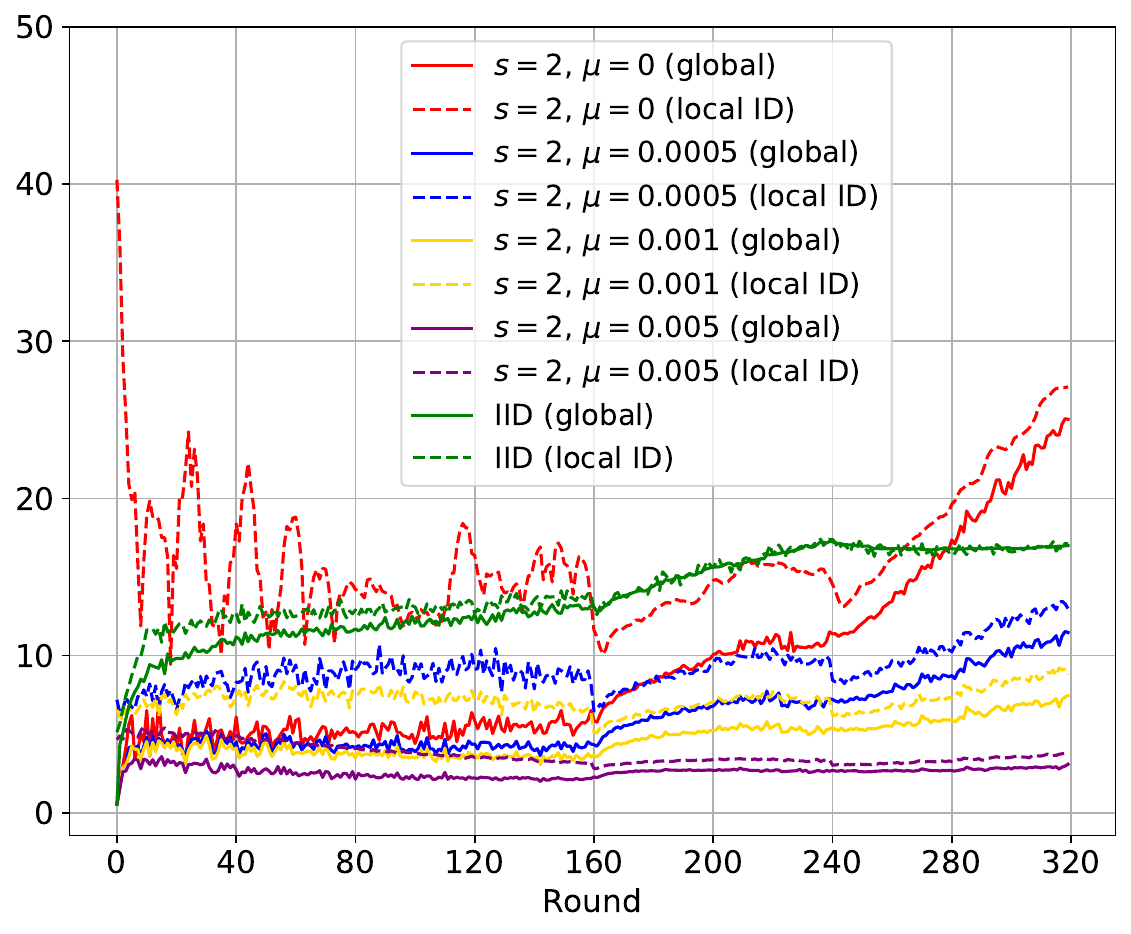}
        \label{fig:fedfr-local-global}
        \vspace{-0.25 in}
        \caption{Feature norm means of local and global models}
    \end{subfigure}
    \caption{Effects of Feature Norm Regularization.}
    \label{fig:fedfr-result}
\end{figure}
\vspace{-0.165 in}
To mitigate feature norm discrepancy between local and global models, especially in scenarios of high heterogeneity that lead to elevated local model feature norms, we introduce an L2 feature norm regularization term with a hyperparameter $\mu$ to the local model training loss. This term is added alongside the cross-entropy loss $\mathcal{L}_{CE}$, and the combined loss $\mathcal{L}_{\mu}$ is formulated as follows:
\begin{equation}
    \mathcal{L}_{\mu}(x;\theta)=\mathcal{L}_{CE}(x;\theta)+\mu\,||f(x;\theta_{ext})||_{2}.
\end{equation}\label{eqn:fedfr}
We apply $\mathcal{L}_{\mu}$ with various hyperparameters $\mu\in\{0, 0.0001, 0.0005, 0.001, 0.005, 0.01, 0.05\}$ across different $s\in\{2,3,5,10\}$ settings, and the results are illustrated in Figure~\ref{fig:fedfr-result}. 

The optimal hyperparameter $\mu$ varies across different heterogeneity settings. For example, in Figure~\ref{fig:fedfr-result} (a) , we observe that for $s=5$ and $s=2$, the optimal $\mu$ values are 0.05 and 0.005, respectively. In Figure~\ref{fig:fedfr-result} (b), we illustrate the feature norm means of local and global models during training under high heterogeneity setting ($s$=2) for different $\mu$ values. 
Notably, at the optimal $\mu$ value of 0.005,  we observe a significant reduction in these discrepancies, aligning more closely with the ideal IID setting. Moreover, each feature norm mean of local and global models are noticeably lower than those in the ideal IID setting. This difference can be attributed to the decrease in the feature norm mean of local model, which subsequently impacts the feature norm mean of the global model. In summary, our findings underscore the performance improvements can be achieved by reducing discrepancies in feature norm means between the global and local models during the training process.

\section{FedFN: Federated Averaging with Feature Normalization Update}\label{sec:method}
In this section, we present Federated Averaging with Feature Normalization (FedFN), which integrates feature normalization (FN) with FedAVG. We also apply the four-factor analysis conducted in Section~\ref{sec:problem} to FedFN. Furthermore, the gradual reduction in weight norm bias within FedFN during training is detailed in Appendix~\ref{app:add_exp}.
\vspace{-0.1 in}
\subsection{FedFN Algorithm}\label{subsec:method}
\vspace{-0.05 in}
We revisit \emph{Federated Averaging with Feature Normalization Update} (FedFN), an extension of FedAVG enriched with Feature Normalization (FN) updates as discussed in \cite{dong2022spherefed}\footnote{SphereFed~\citep{dong2022spherefed} proposes an approach to FL where the classifier is initialized in an orthonormalized manner and kept frozen. Meanwhile, feature normalization is applied to train the local model, utilizing MSE loss. The comparison between FedFN and SphereFed can be found in Appendix~\ref{app:add_exp}.}. 
FN eliminates feature norm bias within the local model by normalizing the feature vector, ensuring that the norm is consistently set to 1 for any input $x$. In FedFN, compared to FedAVG, the FN update modifies the logit vector $z$ of an input $x$, represented as $\hat{z}(x;\theta)=\theta_{cls}\frac{f(x;\theta_{ext})}{||f(x;\theta_{ext})||_{2}}$. Consequently, the gradient of $\theta_\text{cls}$ concerning the cross-entropy loss $\mathcal{L}_{CE}(\cdot)$ for FedFN is expressed as follows:
\vspace{-0.05 in}
\begin{equation*}
\nabla_{\theta_{cls}}\mathcal{L}_{CE}(x;\theta)=\nabla_{\hat{z}(x;\theta)}\mathcal{L}_{CE}(x;\theta)\frac{f(x;\theta_{ext})^{\top}}{||f(x;\theta_{ext})||_{2}}\in \mathds{R}^{C \times d}.
\end{equation*}
\begin{wraptable}[6]{r}{0.45\textwidth}
\centering
\vspace{-0.2in}
\caption{Accuracy of FedAVG and FedFN.}
    \begin{tabular}{c|cc}
    \toprule
     & FedAVG  & FedFN  \\ \midrule 
    $s$=10  & 81.97 & \textbf{83.80}  \\ 
    $s$=2 &  74.24 & \textbf{77.77} \\
    \bottomrule
    \end{tabular}
\label{tab:acc_fedavg_fedfn}
\end{wraptable}
Unlike FedAVG, FedFN scales the gradient of $\theta_{cls}$ by dividing it by the feature vector norm. This scaling significantly impacts the gradient of $\theta_{cls}$ and, consequently, the applied learning rate. As a result of this influence, we conduct  a thorough fine-tuning for the learning rate for the FN update, leading FedFN to adopt a larger initial learning rate of 0.03, compared to the baseline rate of 0.01 in FedAVG.
These learning rates undergo careful selection through an extensive grid search, with detailed findings available in Appendix~\ref{app:grid search}.  Table~\ref{tab:acc_fedavg_fedfn} demonstrates significant accuracy improvement with FedFN compared to FedAVG.
\vspace{-0.1 in}
\subsection{4-Factor Analysis of FedFN}\label{subsec:4-factor-fn}
\begin{figure}[htp]
    \centering
    \vspace{-0.15 in}
    \makebox[\textwidth]{
    \includegraphics[width=1.0\textwidth]{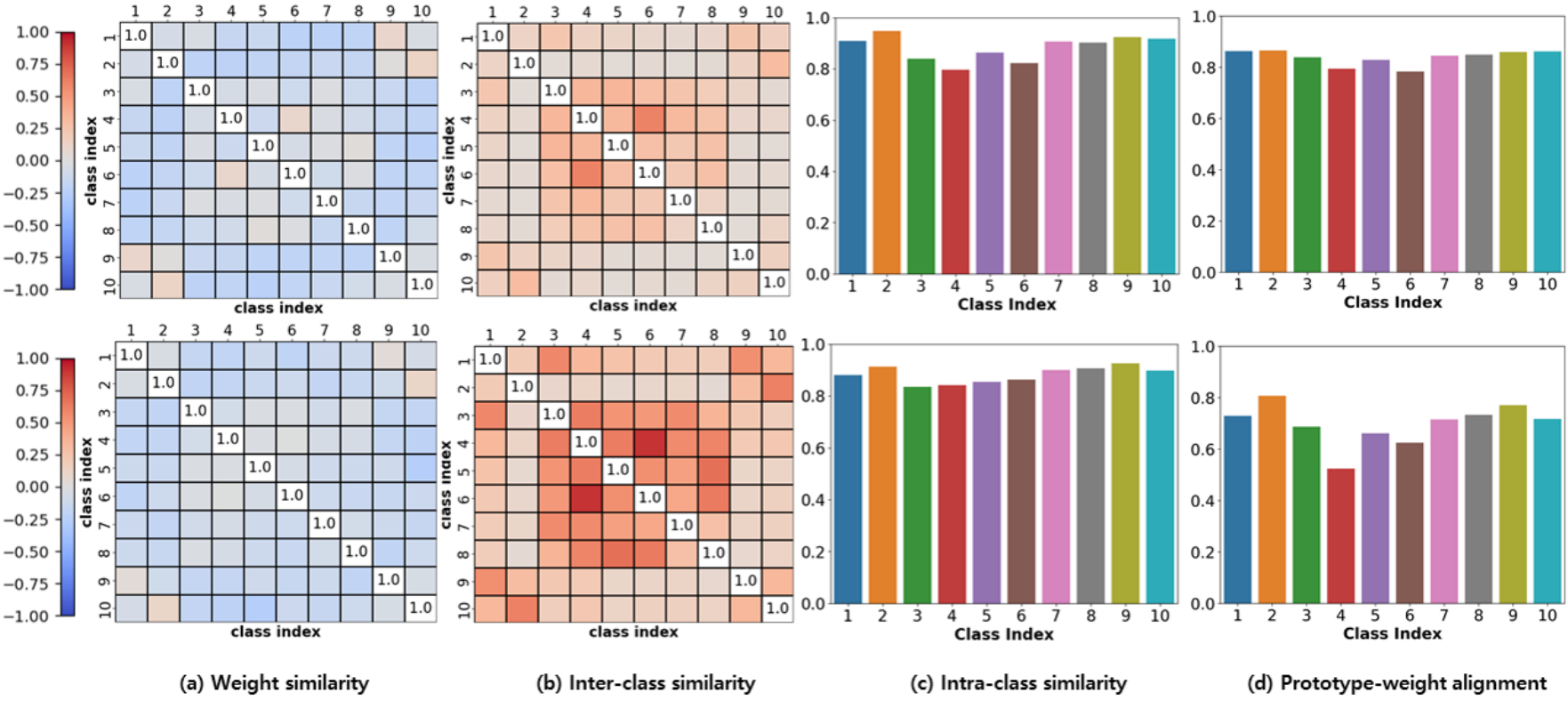}
    }
    \caption{4-Factor Analysis of FedFN; Results for $s$=10 (Top Row) and $s$=2 (Bottom Row).} 
    \label{fig:4_factor_fedFN}
\end{figure}
\vspace{-0.05 in}
We conduct four factor analysis on the improved global model compared to FedAVG, as reported in Table~\ref{tab:acc_fedavg_fedfn}. The results are visualized in Figure \ref{fig:4_factor_fedFN}. Similar to FedAVG, weight similarity remains robust even in high data heterogeneity. On the other hand, FedFN demonstrates significant improvement in inter-class similarity between prototypes compared to FedAVG, especially in the high data heterogeneity. In contrast to FedAVG, FedFN exhibits a slight decrease in intra-class similarity as data heterogeneity increases. This shift likely be attributed to the simultaneous improvement in inter-class similarity within FedFN, enabling finer class separation. While some degradation in prototype-weight alignment is observed in FedFN, FedFN consistently outperforms FedAVG, possibly due to its enhanced inter-class similarity. In summary, the superior inter-class similarity of FedFN contributes to its enhanced performance, resulting in improved discrimination and stability across various data heterogeneity.
\section{Experiment Results}\label{sec:exp}
\vspace{-0.15 in}
In this section, we present the experimental results of FedFN, demonstrating its compatibility with existing FL algorithms. Moreover, we investigate a comparative analysis with the pretrained model To assess the feasibility of implementing FedFN in the foundation models. Additionally, the application of FedFN for the personalized FL can be found in Appendix~\ref{app:add_exp}.
\vspace{-0.1 in}
\subsection{Compatibility of FN with Existing FL Algorithms}\label{subsec:gfl_compatible}
\vspace{-0.1 in}
We assess the compatibility of the FN update module with existing FL algorithms, including FedAVG~\cite{mcmahan2017communication}, Scaffold~\cite{karimireddy2020scaffold}, and FedEXP~\cite{jhunjhunwala2023fedexp} (referred to as 
``Baseline'' algorithms). Additionally, we compare the results with those obtained by applying the BABU module~\citep{oh2021fedbabu}, which freezes the classifier part of the model, to the baseline algorithms. Table~\ref{tab:gfl_acc} summarizes the accuracy comparison for different FL algorithms, including baseline, +BABU, and +FN, on VGG11 for CIFAR-10 and MobileNet for CIFAR-100. For CIFAR-10 and CIFAR-100, we use FN updates with initial learning rates of 0.03 and 0.5, respectively. 
While Scaffold generally demonstrates superior performance, it faces training failures in scenarios with high data heterogeneity, even when applying the BABU module (e.g., $s$=3). 
In contrast, the FN update consistently outperforms all algorithms, including baselines and those with the BABU module, across all heterogeneity settings, demonstrating its superiority.
\vspace{-0.1 in}
\begin{table}[h!]
    \centering
    \vspace{-0.15 in}
    \caption{FL Accuracy Comparison for Baseline, +BABU, and +FN.}
    \small
    \resizebox{0.85\textwidth}{!}{
    \begin{tabular}{cc|cccc|ccc}
      \toprule 
      \multirow{2.5}{*}{Algorithm} & \multirow{2.5}{*}{Module} & \multicolumn{4}{c|}{VGG11 on CIFAR-10} & \multicolumn{3}{c}{MobileNet on CIFAR-100} \\ \cmidrule{3-9}
      & & $s$=2 & $s$=3 & $s$=5 & $s$=10 & $s$=10 & $s$=50 & $s$=100 \\ \midrule
      & Baseline  & 73.62 & 77.70 & 81.62 & 82.13 & 37.25 & 42.90 & 43.36 \\ 
      FedAVG (2017) & + BABU & 73.11 & 76.57 & 81.22 & 81.89 & 43.20 & 39.70 & 39.59 \\ 
      & \textbf{+ FN} & \textbf{76.47} & \textbf{78.66} & \textbf{82.07} & \textbf{83.09} & \textbf{44.67} & \textbf{48.17} & \textbf{49.67}  \\  \midrule
      & Baseline & 77.07 & (\textit{Failed}) & \textbf{84.30} & 84.98 & 41.86 & 43.59 & 41.74 \\
      Scaffold (2020)   & + BABU & 77.17 & (\textit{Failed}) & 83.54 & 84.43 & 46.41 & 41.61 & 42.55 \\ 
       & \textbf{+ FN} & \textbf{77.96} & \textbf{79.24} & \textbf{84.40} & \textbf{85.54} & \textbf{49.42} & \textbf{50.42} & \textbf{52.10}  \\  \midrule
      & Baseline & 73.49 & 77.90 & 81.64 & 82.42 & 36.35 & 41.06 & 42.38\\
      FedEXP (2023) & +BABU & 72.58 & 77.59 & 81.07 & 81.96 & 43.38 & 40.73 & 39.04 \\ 
      & \textbf{+ FN} & \textbf{76.33} & \textbf{78.20} & \textbf{82.41} & \textbf{83.26} & \textbf{45.90} & \textbf{49.10} & \textbf{49.11}  \\  \bottomrule 
    \end{tabular}
    }
    \label{tab:gfl_acc}
\end{table}
\vspace{-0.25 in}
\subsection{Comparative Analysis with Pretrained Model}\label{subsec:gfl_foundation}
\vspace{-0.05 in}
To evaluate the feasibility of applying FedFN within the foundation models, we conducted experiments using a pretrained ResNet18 model on the CIFAR-10 dataset~\citep{yu2023federated, tan2022federated, chen2206importance, nguyen2210begin}. Specifically, we compare the performance of FedAVG, FedBABU, and FedFN when utilizing both pretrained and non-pretrained ResNet18 models. Table~\ref{tab:gfl_foundation} presents the accuracy comparison for these algorithms. Across all experimental settings, FedFN consistently outperforms both FedBABU and FedAVG. Notably, FedAVG and FedBABU, particularly in scenarios with high data heterogeneity, exhibit significant performance declines when pretrained models are employed. In contrast, FedFN consistently improves with the application of pretrained models, utilizing FN updates with an initial learning rate of 0.1.


\begin{table}[htp]
    \centering
    \vspace{-0.15 in}
    \caption{Accuracy Comparison on CIFAR-10 on ResNet18.}
    \small
    \begin{tabular}{c|cccc|cccc}
    \toprule
    \multirow{2.5}{*}{Algorithm} & \multicolumn{4}{c|}{Pretrained=\emph{False}} & \multicolumn{4}{c}{Pretrained=\emph{True}} \\ \cmidrule{2-9} 
                                 & $s$=2 & $s$=3 & $s$=5 & $s$=10 & $s$=2 & $s$=3 & $s$=5 & $s$=10 \\ \midrule
    FedAVG   & 41.50 & 55.31 & 67.64 & 73.39 & 37.87 & 58.31 & 71.75 & 84.50 \\ 
    FedBABU  & 49.21 & 58.44 & 68.84 & 73.82 & 49.78 & 49.61 & 66.46 & 84.22 \\ 
    FedFN    & \textbf{55.17} & \textbf{60.47} & \textbf{77.12} & \textbf{81.26} & \textbf{56.84} & \textbf{76.84} & \textbf{80.02} & \textbf{84.99} \\ \bottomrule                            
    \end{tabular}    \label{tab:gfl_foundation}
\end{table}

\section{Conclusion}\label{sec: discussion}
\vspace{-0.1 in}
In this study, we observe that increasing data heterogeneity leads to larger feature norm disparities between global and local models, which are influenced by feature norm bias within local models. 
We address this issue by introducing feature normalization techniques. Extensive experiments across various FL confirm the superior performance of feature normalization, emphasizing its role in enhancing the feature representations. Our experiments showcase the exceptional performance of FedFN, extending its effectiveness to pretrained ResNet18 models and confirming its applicability to foundational models.

\section*{Acknowledgement}
This work was supported by Institute of Information \& communications Technology Planning \& Evaluation (IITP) grant funded by Korea government (MSIT) [No. 2021-0-00907, Development of Adaptive and Lightweight Edge-Collaborative Analysis Technology for Enabling Proactively Immediate Response and Rapid Learning, 90\%] and [No. 2019-0-00075, Artificial Intelligence Graduate School Program (KAIST), 10\%]. We thank Minchan Jeong for the invaluable discussions and experimental support.

\newpage
\bibliographystyle{plainnat}
\bibliography{ref}

\begin{thebibliography}{43}
\providecommand{\natexlab}[1]{#1}
\providecommand{\url}[1]{\texttt{#1}}
\expandafter\ifx\csname urlstyle\endcsname\relax
  \providecommand{\doi}[1]{doi: #1}\else
  \providecommand{\doi}{doi: \begingroup \urlstyle{rm}\Url}\fi

\bibitem[Arivazhagan et~al.(2019)Arivazhagan, Aggarwal, Singh, and
  Choudhary]{arivazhagan2019federated}
Manoj~Ghuhan Arivazhagan, Vinay Aggarwal, Aaditya~Kumar Singh, and Sunav
  Choudhary.
\newblock Federated learning with personalization layers.
\newblock \emph{arXiv preprint arXiv:1912.00818}, 2019.

\bibitem[Chen et~al.(2022)Chen, Tu, Li, Shen, and Chao]{chen2206importance}
Hong-You Chen, Cheng-Hao Tu, Ziwei Li, Han-Wei Shen, and Wei-Lun Chao.
\newblock On the importance and applicability of pre-training for federated
  learning.
\newblock \emph{URL https://arxiv. org/abs/2206.11488}, 2022.

\bibitem[Collins et~al.(2021)Collins, Hassani, Mokhtari, and
  Shakkottai]{collins2021exploiting}
Liam Collins, Hamed Hassani, Aryan Mokhtari, and Sanjay Shakkottai.
\newblock Exploiting shared representations for personalized federated
  learning.
\newblock In \emph{International conference on machine learning}, pages
  2089--2099. PMLR, 2021.

\bibitem[Dong et~al.(2022)Dong, Zhang, Li, and Kung]{dong2022spherefed}
Xin Dong, Sai~Qian Zhang, Ang Li, and HT~Kung.
\newblock Spherefed: Hyperspherical federated learning.
\newblock In \emph{European Conference on Computer Vision}, pages 165--184.
  Springer, 2022.

\bibitem[Fallah et~al.(2020)Fallah, Mokhtari, and
  Ozdaglar]{fallah2020personalized}
Alireza Fallah, Aryan Mokhtari, and Asuman Ozdaglar.
\newblock Personalized federated learning with theoretical guarantees: A
  model-agnostic meta-learning approach.
\newblock \emph{Advances in Neural Information Processing Systems},
  33:\penalty0 3557--3568, 2020.

\bibitem[Hasnat et~al.(2017)Hasnat, Bohn{\'e}, Milgram, Gentric, and
  Chen]{hasnat2017deepvisage}
Abul Hasnat, Julien Bohn{\'e}, Jonathan Milgram, St{\'e}phane Gentric, and
  Liming Chen.
\newblock Deepvisage: Making face recognition simple yet with powerful
  generalization skills.
\newblock In \emph{Proceedings of the IEEE International Conference on Computer
  Vision Workshops}, pages 1682--1691, 2017.

\bibitem[He et~al.(2016)He, Zhang, Ren, and Sun]{he2016identity}
Kaiming He, Xiangyu Zhang, Shaoqing Ren, and Jian Sun.
\newblock Identity mappings in deep residual networks.
\newblock In \emph{Computer Vision--ECCV 2016: 14th European Conference,
  Amsterdam, The Netherlands, October 11--14, 2016, Proceedings, Part IV 14},
  pages 630--645. Springer, 2016.

\bibitem[Howard et~al.(2017)Howard, Zhu, Chen, Kalenichenko, Wang, Weyand,
  Andreetto, and Adam]{howard2017mobilenets}
Andrew~G Howard, Menglong Zhu, Bo~Chen, Dmitry Kalenichenko, Weijun Wang,
  Tobias Weyand, Marco Andreetto, and Hartwig Adam.
\newblock Mobilenets: Efficient convolutional neural networks for mobile vision
  applications.
\newblock \emph{arXiv preprint arXiv:1704.04861}, 2017.

\bibitem[Huang et~al.(2023)Huang, Xie, Yang, Wang, Lin, and
  Cai]{huang2023neural}
Chenxi Huang, Liang Xie, Yibo Yang, Wenxiao Wang, Binbin Lin, and Deng Cai.
\newblock Neural collapse inspired federated learning with non-iid data, 2023.

\bibitem[Jhunjhunwala et~al.(2023)Jhunjhunwala, Wang, and
  Joshi]{jhunjhunwala2023fedexp}
Divyansh Jhunjhunwala, Shiqiang Wang, and Gauri Joshi.
\newblock Fedexp: Speeding up federated averaging via extrapolation.
\newblock \emph{arXiv preprint arXiv:2301.09604}, 2023.

\bibitem[Kang et~al.(2019)Kang, Xie, Rohrbach, Yan, Gordo, Feng, and
  Kalantidis]{kang2019decoupling}
Bingyi Kang, Saining Xie, Marcus Rohrbach, Zhicheng Yan, Albert Gordo, Jiashi
  Feng, and Yannis Kalantidis.
\newblock Decoupling representation and classifier for long-tailed recognition.
\newblock \emph{arXiv preprint arXiv:1910.09217}, 2019.

\bibitem[Karimireddy et~al.(2020)Karimireddy, Kale, Mohri, Reddi, Stich, and
  Suresh]{karimireddy2020scaffold}
Sai~Praneeth Karimireddy, Satyen Kale, Mehryar Mohri, Sashank Reddi, Sebastian
  Stich, and Ananda~Theertha Suresh.
\newblock Scaffold: Stochastic controlled averaging for federated learning.
\newblock In \emph{International conference on machine learning}, pages
  5132--5143. PMLR, 2020.

\bibitem[Khosla et~al.(2020)Khosla, Teterwak, Wang, Sarna, Tian, Isola,
  Maschinot, Liu, and Krishnan]{khosla2020supervised}
Prannay Khosla, Piotr Teterwak, Chen Wang, Aaron Sarna, Yonglong Tian, Phillip
  Isola, Aaron Maschinot, Ce~Liu, and Dilip Krishnan.
\newblock Supervised contrastive learning.
\newblock \emph{Advances in neural information processing systems},
  33:\penalty0 18661--18673, 2020.

\bibitem[Krizhevsky et~al.(2009)Krizhevsky, Nair, and
  Hinton]{krizhevsky2009cifar}
Alex Krizhevsky, Vinod Nair, and Geoffrey Hinton.
\newblock Cifar-10 and cifar-100 datasets.
\newblock \emph{URl: https://www. cs. toronto. edu/kriz/cifar. html},
  6\penalty0 (1):\penalty0 1, 2009.

\bibitem[Li et~al.(2021{\natexlab{a}})Li, He, and Song]{li2021model}
Qinbin Li, Bingsheng He, and Dawn Song.
\newblock Model-contrastive federated learning.
\newblock In \emph{Proceedings of the IEEE/CVF conference on computer vision
  and pattern recognition}, pages 10713--10722, 2021{\natexlab{a}}.

\bibitem[Li et~al.(2020)Li, Sahu, Zaheer, Sanjabi, Talwalkar, and
  Smith]{li2020federated}
Tian Li, Anit~Kumar Sahu, Manzil Zaheer, Maziar Sanjabi, Ameet Talwalkar, and
  Virginia Smith.
\newblock Federated optimization in heterogeneous networks.
\newblock \emph{Proceedings of Machine learning and systems}, 2:\penalty0
  429--450, 2020.

\bibitem[Li et~al.(2021{\natexlab{b}})Li, Hu, Beirami, and Smith]{li2021ditto}
Tian Li, Shengyuan Hu, Ahmad Beirami, and Virginia Smith.
\newblock Ditto: Fair and robust federated learning through personalization.
\newblock In \emph{International Conference on Machine Learning}, pages
  6357--6368. PMLR, 2021{\natexlab{b}}.

\bibitem[Li et~al.(2019)Li, Huang, Yang, Wang, and Zhang]{li2019convergence}
Xiang Li, Kaixuan Huang, Wenhao Yang, Shusen Wang, and Zhihua Zhang.
\newblock On the convergence of fedavg on non-iid data.
\newblock \emph{arXiv preprint arXiv:1907.02189}, 2019.

\bibitem[Li and Zhan(2021)]{li2021fedrs}
Xin-Chun Li and De-Chuan Zhan.
\newblock Fedrs: Federated learning with restricted softmax for label
  distribution non-iid data.
\newblock In \emph{Proceedings of the 27th ACM SIGKDD Conference on Knowledge
  Discovery \& Data Mining}, pages 995--1005, 2021.

\bibitem[Li et~al.(2022)Li, Xu, Song, Li, Li, Shao, and Zhan]{li2022federated}
Xin-Chun Li, Yi-Chu Xu, Shaoming Song, Bingshuai Li, Yinchuan Li, Yunfeng Shao,
  and De-Chuan Zhan.
\newblock Federated learning with position-aware neurons.
\newblock In \emph{Proceedings of the IEEE/CVF Conference on Computer Vision
  and Pattern Recognition}, pages 10082--10091, 2022.

\bibitem[Li et~al.(2023)Li, Shang, He, Lin, and Wu]{li2023no}
Zexi Li, Xinyi Shang, Rui He, Tao Lin, and Chao Wu.
\newblock No fear of classifier biases: Neural collapse inspired federated
  learning with synthetic and fixed classifier.
\newblock \emph{arXiv preprint arXiv:2303.10058}, 2023.

\bibitem[Lin et~al.(2020)Lin, Kong, Stich, and Jaggi]{lin2020ensemble}
Tao Lin, Lingjing Kong, Sebastian~U Stich, and Martin Jaggi.
\newblock Ensemble distillation for robust model fusion in federated learning.
\newblock \emph{Advances in Neural Information Processing Systems},
  33:\penalty0 2351--2363, 2020.

\bibitem[Luo et~al.(2021)Luo, Chen, Hu, Zhang, Liang, and Feng]{luo2021no}
Mi~Luo, Fei Chen, Dapeng Hu, Yifan Zhang, Jian Liang, and Jiashi Feng.
\newblock No fear of heterogeneity: Classifier calibration for federated
  learning with non-iid data.
\newblock \emph{Advances in Neural Information Processing Systems},
  34:\penalty0 5972--5984, 2021.

\bibitem[McMahan et~al.(2017)McMahan, Moore, Ramage, Hampson, and
  y~Arcas]{mcmahan2017communication}
Brendan McMahan, Eider Moore, Daniel Ramage, Seth Hampson, and Blaise~Aguera
  y~Arcas.
\newblock Communication-efficient learning of deep networks from decentralized
  data.
\newblock In \emph{Artificial intelligence and statistics}, pages 1273--1282.
  PMLR, 2017.

\bibitem[Mettes et~al.(2019)Mettes, Van~der Pol, and
  Snoek]{mettes2019hyperspherical}
Pascal Mettes, Elise Van~der Pol, and Cees Snoek.
\newblock Hyperspherical prototype networks.
\newblock \emph{Advances in neural information processing systems}, 32, 2019.

\bibitem[Nguyen et~al.(2022)Nguyen, Wang, Malik, Sanjabi, and
  Rabbat]{nguyen2210begin}
J~Nguyen, J~Wang, K~Malik, M~Sanjabi, and M~Rabbat.
\newblock Where to begin? on the impact of pre-training and initialization in
  federated learning.
\newblock \emph{arXiv preprint arXiv:2210.08090}, 2022.

\bibitem[Oh et~al.(2021)Oh, Kim, and Yun]{oh2021fedbabu}
Jaehoon Oh, Sangmook Kim, and Se-Young Yun.
\newblock Fedbabu: Towards enhanced representation for federated image
  classification.
\newblock \emph{arXiv preprint arXiv:2106.06042}, 2021.

\bibitem[Papyan et~al.(2020)Papyan, Han, and Donoho]{papyan2020prevalence}
Vardan Papyan, XY~Han, and David~L Donoho.
\newblock Prevalence of neural collapse during the terminal phase of deep
  learning training.
\newblock \emph{Proceedings of the National Academy of Sciences}, 117\penalty0
  (40):\penalty0 24652--24663, 2020.

\bibitem[Savvides et~al.(2021)Savvides, Pal, and Zheng]{savvides2021convex}
Marios Savvides, Dipan~Kumar Pal, and Yutong Zheng.
\newblock Convex feature normalization for face recognition, February~4 2021.
\newblock US Patent App. 16/968,040.

\bibitem[Shi et~al.(2022)Shi, Liang, Zhang, Tan, and Bai]{shi2022towards}
Yujun Shi, Jian Liang, Wenqing Zhang, Vincent~YF Tan, and Song Bai.
\newblock Towards understanding and mitigating dimensional collapse in
  heterogeneous federated learning.
\newblock \emph{arXiv preprint arXiv:2210.00226}, 2022.

\bibitem[Simonyan and Zisserman(2014)]{simonyan2014very}
Karen Simonyan and Andrew Zisserman.
\newblock Very deep convolutional networks for large-scale image recognition.
\newblock \emph{arXiv preprint arXiv:1409.1556}, 2014.

\bibitem[Tan et~al.(2022{\natexlab{a}})Tan, Long, Liu, Zhou, Lu, Jiang, and
  Zhang]{tan2022fedproto}
Yue Tan, Guodong Long, Lu~Liu, Tianyi Zhou, Qinghua Lu, Jing Jiang, and Chengqi
  Zhang.
\newblock Fedproto: Federated prototype learning across heterogeneous clients.
\newblock In \emph{Proceedings of the AAAI Conference on Artificial
  Intelligence}, pages 8432--8440, 2022{\natexlab{a}}.

\bibitem[Tan et~al.(2022{\natexlab{b}})Tan, Long, Ma, Liu, Zhou, and
  Jiang]{tan2022federated}
Yue Tan, Guodong Long, Jie Ma, Lu~Liu, Tianyi Zhou, and Jing Jiang.
\newblock Federated learning from pre-trained models: A contrastive learning
  approach.
\newblock \emph{Advances in Neural Information Processing Systems},
  35:\penalty0 19332--19344, 2022{\natexlab{b}}.

\bibitem[Wang et~al.(2018)Wang, Cheng, Liu, and Liu]{wang2018additive}
Feng Wang, Jian Cheng, Weiyang Liu, and Haijun Liu.
\newblock Additive margin softmax for face verification.
\newblock \emph{IEEE Signal Processing Letters}, 25\penalty0 (7):\penalty0
  926--930, 2018.

\bibitem[Wang et~al.(2020{\natexlab{a}})Wang, Yurochkin, Sun, Papailiopoulos,
  and Khazaeni]{wang2020federated}
Hongyi Wang, Mikhail Yurochkin, Yuekai Sun, Dimitris Papailiopoulos, and
  Yasaman Khazaeni.
\newblock Federated learning with matched averaging.
\newblock \emph{arXiv preprint arXiv:2002.06440}, 2020{\natexlab{a}}.

\bibitem[Wang et~al.(2020{\natexlab{b}})Wang, Liu, Liang, Joshi, and
  Poor]{wang2020tackling}
Jianyu Wang, Qinghua Liu, Hao Liang, Gauri Joshi, and H~Vincent Poor.
\newblock Tackling the objective inconsistency problem in heterogeneous
  federated optimization.
\newblock \emph{Advances in neural information processing systems},
  33:\penalty0 7611--7623, 2020{\natexlab{b}}.

\bibitem[Wu et~al.(2019)Wu, Chen, Wang, Ye, Liu, Guo, and Fu]{wu2019large}
Yue Wu, Yinpeng Chen, Lijuan Wang, Yuancheng Ye, Zicheng Liu, Yandong Guo, and
  Yun Fu.
\newblock Large scale incremental learning.
\newblock In \emph{Proceedings of the IEEE/CVF conference on computer vision
  and pattern recognition}, pages 374--382, 2019.

\bibitem[Xu et~al.(2023)Xu, Tong, and Huang]{xu2023personalized}
Jian Xu, Xinyi Tong, and Shao-Lun Huang.
\newblock Personalized federated learning with feature alignment and classifier
  collaboration.
\newblock \emph{arXiv preprint arXiv:2306.11867}, 2023.

\bibitem[Yu et~al.(2021)Yu, Zhang, Qin, Xu, Wang, Liu, Tian, and
  Chen]{yu2021fed2}
Fuxun Yu, Weishan Zhang, Zhuwei Qin, Zirui Xu, Di~Wang, Chenchen Liu, Zhi Tian,
  and Xiang Chen.
\newblock Fed2: Feature-aligned federated learning.
\newblock In \emph{Proceedings of the 27th ACM SIGKDD conference on knowledge
  discovery \& data mining}, pages 2066--2074, 2021.

\bibitem[Yu et~al.(2020)Yu, Zhang, Deng, Yuan, Jia, and Chen]{yu2020devil}
Haiyang Yu, Ningyu Zhang, Shumin Deng, Zonggang Yuan, Yantao Jia, and Huajun
  Chen.
\newblock The devil is the classifier: Investigating long tail relation
  classification with decoupling analysis.
\newblock \emph{arXiv preprint arXiv:2009.07022}, 2020.

\bibitem[Yu et~al.(2023)Yu, Mu{\~n}oz, and Jannesari]{yu2023federated}
Sixing Yu, J~Pablo Mu{\~n}oz, and Ali Jannesari.
\newblock Federated foundation models: Privacy-preserving and collaborative
  learning for large models.
\newblock \emph{arXiv preprint arXiv:2305.11414}, 2023.

\bibitem[Zhao et~al.(2020)Zhao, Xiao, Gan, Zhang, and Xia]{zhao2020maintaining}
Bowen Zhao, Xi~Xiao, Guojun Gan, Bin Zhang, and Shu-Tao Xia.
\newblock Maintaining discrimination and fairness in class incremental
  learning.
\newblock In \emph{Proceedings of the IEEE/CVF conference on computer vision
  and pattern recognition}, pages 13208--13217, 2020.

\bibitem[Zhao et~al.(2018)Zhao, Li, Lai, Suda, Civin, and
  Chandra]{zhao2018federated}
Yue Zhao, Meng Li, Liangzhen Lai, Naveen Suda, Damon Civin, and Vikas Chandra.
\newblock Federated learning with non-iid data.
\newblock \emph{arXiv preprint arXiv:1806.00582}, 2018.

\end{thebibliography}

\newpage
\appendix
\appendix
\pagebreak

\begin{center}
    \noindent\\
    \vspace{0.2in}
    \textbf{\LARGE -- Appendix --}\\
    \vspace{0.1in}
    \textbf{\LARGE FedFN: Feature Normalization for Alleviating Data Heterogeneity in Federated Learning}\\
    \vspace{0.1in}
    \noindent
\end{center}

\setcounter{thm}{0}
\setcounter{equation}{0}
\setcounter{table}{0}
\setcounter{figure}{0}

\section{Related Work}
\label{sec:related}

\subsection{Federated Learing}
\textbf{Global Federated Learning}
Global Federated Learning (GFL) aims to enhance the performance of a single global model across decentralized clients by addressing data heterogeneity arising from diverse user behaviors. Researchers have explored various methodologies within GFL to create robust models for diverse devices and data sources. These approaches include client drift mitigation~\citep{li2020federated,karimireddy2020scaffold,jhunjhunwala2023fedexp}, aggregation schemes to improve model fusion mechanisms at the server~\citep{wang2020federated, wang2020tackling}, and data sharing techniques introducing public datasets or synthesized data to achieve a more balanced data distribution~\citep{zhao2018federated, lin2020ensemble, luo2021no}.

\noindent{\textbf{Personalized Federated Learning}}
Personalized Federated Learning (PFL) focuses on training personalized models for individual clients, adapting to their specific data distributions and tasks. PFL methodologies include decoupling methods that separate the feature extractor and classifier during communication, enabling unique updates for the data distribution of each client~\citep{arivazhagan2019federated, collins2021exploiting, oh2021fedbabu}, modifying local loss functions to improve task performance~\citep{fallah2020personalized, li2021ditto}, and utilizing prototype communication techniques~\citep{tan2022fedproto, xu2023personalized}.

\subsection{Partial Model Updates in FL}
\textbf{Debiasing classifier in FL}
Efforts to address data heterogeneity in both GFL and PFL domains have explored differential updates within the model parameters, with a particular focus on the classifier part. For instance, \citet{luo2021no} propose classifier post-calibration with virtual features to tackle a notable bias among classifiers of different local models. \citet{li2021fedrs} introduce the restricted softmax loss for local updates to prevent classifiers from becoming inaccurate when updating for missing classes. Additionally, certain studies~\citep{oh2021fedbabu, dong2022spherefed, li2023no, huang2023neural} suggest the use of fixed classifiers constructed from orthogonal basis vectors during training.

\noindent\textbf{Feature Enhancement in FL} Recent research~\citep{yu2021fed2, li2021model, li2022federated, shi2022towards} in the context of GFL has focused on aligning feature representations among local models. For instance, \citet{li2021model} introduce the contrastive loss during local iteration to improve feature alignment. Additionally, \citet{shi2022towards} highlights the issue of data heterogeneity leading to severe dimensional collapse in the global model, resulting in representations tending towards lower dimensions. To address this, they propose using a regularization term during local training to mitigate the issue and improve feature alignment. Moreover, some studies ~\citep{tan2022fedproto, xu2023personalized} have explored communication strategies involving feature prototypes to further enhance feature alignment in the context of PFL.

\subsection{Feature Normalized Model in DL}
Feature normalized model has been widely adapted in various fields of deep learning (DL), including face recognition~\citep{wang2018additive, savvides2021convex}, regression~\citep{mettes2019hyperspherical}, and federated learning~\citep{dong2022spherefed}, with the aim of enhancing the discriminative power of features. In the several studies~\citep{wang2018additive, dong2022spherefed}, both the feature vectors and classifier weights are normalized to enforce cosine-similarity element logits, restricting the values of each element in the logit vector.

\newpage
\section{Preliminaries}
\label{app:prelim}
In the upcoming section, we elucidate the concept of FL, followed by an in-depth discussion of the experimental setup, encompassing dataset descriptions, model specifications, hyperparameter configurations during FL, and detailed description four factor analysis conducted in main paragraph. For clarity and convenience, we present a concise overview of key notations in Table \ref{tab:main_notation_summary}, facilitating comprehension of the paper.  

\begin{table}[htp]
\centering
\caption{Main Notations Throughout the Paper.}
\resizebox{0.8\textwidth}{!}{
\begin{tabular}{ll}

\hline
\textbf{Indices} & \\ 
$c\in[C]$  & index for a class   \\
$r\in[R]$ & index for FL round\\
$n\in[N]$ & index for a client   \\ 

\hline
\textbf{Dataset}&  \\
$(x,y)$ & (input image , true class label of $x$) \\
$D_{train}, D_{test}$ & total train and test dataset \\

$D_{train}^n, D_{test}^n$ & train and test dataset of client $n$\\

$D(c)$ & collection of dataset $D$ with the label $c$\\

\hline
\textbf{Parameters} & \\ 
$\theta:=(\theta_{ext}, \theta_{cls})$  & model parameter   \\
$\theta_{ext}$ & feature extractor part of $\theta$   \\ 
$\theta_{cls}\in \mathds{R}^{C\times d}$ & classifier part of $\theta$ \\
$\theta_{cls,i}, i\in[C]$ & i-th row vector of $\theta_{cls}$ \\
\hline

\textbf{Model Forward}&  \\
$p(x;\theta)\in\mathds{R}^{C}$& softmax probability of input $x$ \\ 
$p_i(x;\theta), i\in[C]$& i-th element of $p(x;\theta)$ \\ 
$\mathcal{L}_{CE}(x;\theta):=-\log p_{y}(x;\theta)$& cross entropy loss of input $x$\\
$f(x;\theta_{ext})\in\mathds{R}^{d}$ & feature vector of input $x$  \\ 
$\hat{f}(x;\theta_{ext}):=f(x;\theta_{ext})/||f(x;\theta_{ext})||_{2}$ & normalized feature vector of input $x$  \\
$z(x;\theta):=\theta_{cls}\,f(x;\theta_{ext})\in\mathds{R}^{C}$& logit vector of input $x$  \\ 
$z_{i}(x;\theta), i\in[C]$& i-th element of $z(x;\theta)$  \\ 
\hline

\hline
\textbf{Prototype  of label c}&  \\
$f(D(c);\theta_{ext}):=\frac{1}{|D(c)|}\sum_{(x,y)\in D(c)}f(x;\theta_{ext})$ & prototype of $D(c)$ \\
$\hat{f}(D(c);\theta_{ext}):=\frac{1}{|D(c)|}\sum_{(x,y)\in D(c)}\hat{f}(x;\theta_{ext})$ & prototype from the normalized features \\
\hline

\end{tabular}
}
\label{tab:main_notation_summary}\end{table}

\subsection{FL Procedure}
In FL, we aim to train a robust image classification model on the central server while preserving the privacy and security of individual client data. The procedure involves communication over $R$ rounds. In each round $r \in [R]$, a random subset of clients $S_r\subset [N]$ is selected from the client pool. These selected clients receive the current global model parameters $\theta^{r-1}$ from the central server. Subsequently, each client $n \in S_r$ performs local updates on its local dataset $D_{train}^n$ for $E$ epochs using a batch size of $B$. The updated model parameters for client $n$ are denoted as ${\theta}^{r,n}$. Afterward, the central server updates the global model parameter $\theta^r$ as a result of convex combination based on the ${\theta}^{r,n}$~\cite{li2019convergence}. This collaborative approach iteratively refines the image classification model on the central server over the $R$ rounds, resulting in a robust model that performs well on the entire test dataset $D_{test}$ while preserving individual client data privacy.

\subsection{Experimental Setup}

\textbf{Datasets and Models}  To simulate a realistic federated learning scenario involving 100 clients, we conduct extensive studies on two widely-used datasets: CIFAR-10 and CIFAR-100~\citep{krizhevsky2009cifar}. For CIFAR-10, we employ the VGG11~\citep{simonyan2014very} model, while for CIFAR-100, the MobileNet~\citep{howard2017mobilenets} model is chosen. The training data is distributed among 100 clients using two distinct Non-IID partition strategies:

\begin{itemize}
\item \textbf{Sharding}~\citep{mcmahan2017communication, oh2021fedbabu}: We meticulously organize the data by label and divide it into non-overlapping shards of equal size. Each shard encompasses $\frac{|D_{train}|}{100\times s}$ samples of the same class, and $s$ denotes the number of shards per client. This results in each client having access to a maximum of $s$ different classes. As we decrease the number of shards per user $s$, the level of data heterogeneity among clients increases. For CIFAR-10, we explore various $s$ values, such as $s\in\{2, 3, 5, 10\}$, while for CIFAR-100, we experiment with $s \in \{10, 50, 100\}$.

\item \textbf{Latent Dirichlet Allocation (LDA)}~\citep{luo2021no, wang2020federated}: We utilize the LDA technique to sample a probability vector $p_c = (p_{c,1}, p_{c,2}, \cdots, p_{c,100}) \sim Dir(\alpha)$ and allocate a proportion $p_{c,k}$ of instances of class $c \in [C]$ to each client $k \in [100]$, where $Dir(\alpha)$ represents the Dirichlet distribution with the concentration parameter $\alpha$. The parameter $\alpha$ controls the strength of data heterogeneity, where smaller values lead to stronger heterogeneity among clients. For both CIFAR-10 and CIFAR-100, we conduct experiments with various $\alpha$ values, such as $\alpha \in \{0.1, 0.3, 0.5, 1.0\}$.

\end{itemize}

\noindent\textbf{Hyperparameter Search:} To optimize the hyperparameters for federated learning, we conduct grid searches for the initial learning rate on both CIFAR-10 and CIFAR-100. For CIFAR-10, we explore learning rates in the range of $\{0.01, 0.03, 0.05, 0.1\}$, while for CIFAR-100, we consider learning rates of $\{0.1, 0.3, 0.5, 1.0\}$. Additionally, we perform grid searches to determine the optimal number of local epochs, evaluating values in the set $\{1, 5, 10, 15, 20\}$ for both datasets. The optimal number of local epochs is found to be 15 for CIFAR-10 and 5 for CIFAR-100. In cases where specific values are not mentioned, we use default initial learning rates of 0.01 for CIFAR-10 and 0.1 for CIFAR-100. 

\noindent\textbf{Implementation Details} All experiments are conducted for 320 rounds to thoroughly assess the performance and convergence behavior of the models. To ensure convergence during training, we decay the learning rate by 0.1 at half and three-quarters of the federated learning rounds. Additionally, we utilize random horizontal flipping as a data augmentation technique throughout the training process. Table~\ref{tab:gfl_acc} in Section~\ref{sec:exp} and Table~\ref{tab:gfl_acc_all} in Appendix~\ref{app:add_exp} are constructed using the code structure from \texttt{https://github.com/Lee-Gihun/FedNTD}, while the rest of the implementations are based on \texttt{https://github.com/jhoon-oh/FedBABU}.

\subsection{4-Factor Analysis}
\label{app:4_factor_analysis}

We introduce four factors, extensively studied~\citep{papyan2020prevalence,kang2019decoupling} and aim to identify which of them have a high negative impact in the presence of data heterogeneity.

\begin{enumerate}[label=\textbf{(\roman*)}]
\item \textbf{Weight similarity:} Measuring the similarity or dissimilarity among classifiers across classes, this factor computes the cosine similarity between their normalized weight vectors, resulting in a symmetric matrix. The detail form is :
\begin{equation*}
N(\theta_{cls})^{\top}N(\theta_{cls})\in [-1,1]^{C\times C}, \text{where}\,\, N(\theta_{cls})=\left[\frac{\theta_{cls,1}^\top}{||\theta_{cls,1}||_{2}}|\cdots ,\frac{\theta_{cls,C}^\top}{||\theta_{cls,C}||_{2}}\right]\in\mathds{R}^{d\times C}.
\end{equation*}
Lower weight similarity is preferred, indicating distinct classifiers for each class and better discrimination.
\item \textbf{Inter-class similarity:} This factor delves into the relationships between feature prototypes representing different classes, represented by a symmetric matrix $f(D_{test};\theta)^{\top}f(D_{test};\theta)\in [-1,1]^{C\times C}$. Here, $f(D_{test};\theta)$  is represented as
\begin{equation*}
    \left[f(D_{test}(1);\theta)|\cdots|f(D_{test}(C);\theta)\right]\in \mathds{R}^{d\times C}.
\end{equation*}

Lower inter-class similarity is desired, representing distinguishable feature vectors for different classes.    
\item \textbf{Intra-class similarity:} This factor provides insights into the diversity or similarity of representations within individual classes. We achieve this by calculating the cosine similarity between feature vectors of test prototypes belonging to the same class.  For each class $c\in [C]$, it is evaluated by:
\begin{equation*}
    \frac{1}{|D_{test}(c)|}\sum_{(x,y)\in D_{test}(c)} \frac{f(x;\theta_{ext})^{\top}}{||f(x;\theta_{ext})||_2}\frac{f(D_{test}(c);\theta_{ext})}{||f(D_{test}(c);\theta_{ext})||_{2}}\in\mathds{R}.
\end{equation*}
Higher intra-class similarity is preferred, indicating that feature vectors belonging to the same class are closer to each other, thereby improving class representation and classification performance.
\item \textbf{Prototype-weight alignment:} Assessing this factor reveals the degree of alignment between the classifier and
prototypes for each class, according to their internal
product. High alignment signifies a strong match, while
low alignment indicates a potential mismatch or poor fit. For each class $c\in [C]$, it is evaluated by inner product of $\theta_{cls,c}$ and $f(D_{test}(c);\theta)$. This factor is generally not a primary concern but may be correlated with weight similarity and inter-class similarity.

\end{enumerate}

\section{Grid Search Result}
\label{app:grid search}
To optimize the hyperparameters for FL, we conduct grid searches for the initial learning rate on both CIFAR-10 and CIFAR-100. For CIFAR-10, we explore learning rate $\eta$ in the range of $\{0.01, 0.03, 0.05, 0.1\}$, while for CIFAR-100, we consider $\eta$ of $\{0.1, 0.3, 0.5, 1.0\}$. Additionally, we performe grid searches to determine the optimal number of local epochs $E$, evaluating values in the set $\{1, 5, 10, 15, 20\}$ for both datasets. The optimal number of $E$ is found to be 15 for CIFAR-10 and 5 for CIFAR-100. Unless otherwise specified, the values determined through grid search for the hyperparameter $\eta$ are as follows: default initial learning rates are 0.01 for CIFAR-10 and 0.1 for CIFAR-100.
\subsection{FedBABU vs FedAVG vs FedFN}
We conduct grid searches for FedAVG~\cite{mcmahan2017communication}, FedBABU~\cite{oh2021fedbabu}, and FedFN. Table ~\ref{apptab:grid_vgg} to ~\ref{apptab:grid_resnet_pretrained} present the grid search results for VGG on CIFAR-10, MobileNet on CIFAR-100, ResNet18 on CIFAR-10, and Pretrained ResNet18 on CIFAR-10, respectively. In these tables, we abbreviate FedAVG, FedBABU, and FedFN as AVG, BABU, and FN, respectively, and indicate cases where the final global model fails to converge during training with a ``-" in the respective table cells. The determined optimal hyperparameter $\eta$ values for FedFN are as follows: 0.03, 0.5, 0.1, and 0.1 for Table ~\ref{apptab:grid_vgg} to ~\ref{apptab:grid_resnet_pretrained}, respectively.
\begin{table}[htp]
    \centering
    \caption{Grid Search Results for VGG11 on CIFAR-10.}  
    \resizebox{\textwidth}{!}{
        \begin{tabular}{c|ccc|ccc|ccc|ccc|ccc}
        \toprule
        \multirow{2.5}{*} {\emph{$\eta$=0.01}} & \multicolumn{3}{c|}{$E$=1}                      & \multicolumn{3}{c|}{$E$=5} & \multicolumn{3}{c|}{$E$=10} & \multicolumn{3}{c|}{$E$=15} & \multicolumn{3}{c}{$E$=20}  \\ \cmidrule{2-16}         
        & BABU & AVG & FN & BABU & AVG & FN & BABU & AVG & FN & BABU & AVG & FN & BABU & AVG & FN \\ \midrule  
        $s$=10 & 49.51 & 52.98  & 52.02 & 82.43 & 82.68 & 82.82 & 82.56 & 83.03        & 82.7  & \textbf{82.16} & \textbf{81.97} & 82.11 & 81.82 & 81.66 & 82.42\\ 
        
        $s$=5 & 42.47 & 46.28 & 50.26 & 79.39 & 80.42 & 79.76 & 81.13 & 81.36 & 82.09 & \textbf{81.04} & \textbf{81.08} & 81.74 & 80.82 & 81.35 & 81.3   \\ 
        
        $s$=3 & 29.45 & 38.32 & 48.24 & 71.64 & 73.83 & 72.94 & 78.39 & 77.39 & 77.49 & \textbf{77.73} & \textbf{77.29}        & 78.37 & 78.00 & 78.03 & 78.48   \\ 
        
        $s$=2 & 24.21 & 30.61 & 46.24 & 56.37 & 63.78 & 62.5 & 73.31 & 73.24 & 73.79  & \textbf{75.05} & \textbf{74.24} & 75.34 & 75.25 & 74.98 & 76.33   \\ \midrule
            
        \multirow{2.5}{*} {\emph{$\eta$=0.03}} & \multicolumn{3}{c|}{$E$=1}                      & \multicolumn{3}{c|}{$E$=5} & \multicolumn{3}{c|}{$E$=10} & \multicolumn{3}{c|}{$E$=15} & \multicolumn{3}{c}{$E$=20}  \\ \cmidrule{2-16}         
        & BABU & AVG & FN & BABU & AVG & FN & BABU & AVG & FN & BABU & AVG & FN & BABU & AVG & FN \\ \midrule  
        $s$=10 & 63.23   & 66.81 & 64.32 & 84.23 & 84.32 & 83.76 & 84.10 & 84.13 & 83.80 & 83.61 & 84.38 & \textbf{83.80} & 83.82 & 83.42 & 83.68   \\ 
        
        $s$=5 & 54.86 & 59.21 & 57.94 & 82.14 & 82.56 & 81.43 & 82.19 & 82.39 & 82.34  & 83.17 & 82.92 & \textbf{82.43} & 82.49 & 82.66 & 82.79   \\ 
        
        $s$=3 & 34.12& 49.90 & 52.11 & 75.71 & - & 74.01 & - & - & 78.51 & - & - & \textbf{78.93} & - & - & 78.77 \\ 
        
        $s$=2 & - & 34.80 & 46.54 & 50.80 & 62.80 & 66.60 & 74.27 & 75.08 & 76.26  & 74.57 & - & \textbf{77.77} & 77.36 & - & 77.67   \\ \midrule
        
        \multirow{2.5}{*} {\emph{$\eta$=0.05}} & \multicolumn{3}{c|}{$E$=1}                      & \multicolumn{3}{c|}{$E$=5} & \multicolumn{3}{c|}{$E$=10} & \multicolumn{3}{c|}{$E$=15} & \multicolumn{3}{c}{$E$=20}  \\ \cmidrule{2-16}         
        & BABU & AVG & FN & BABU & AVG & FN & BABU & AVG & FN & BABU & AVG & FN & BABU & AVG & FN \\ \midrule    
        $s$=10 & 69.58 & 70.99 & 68.22 & 84.49 & 84.51 & 83.60 & 84.22 & 84.30 & 83.35 & 84.19 & 84.18 & 83.81 & 83.94 & - & 82.92 \\ 
        
        $s$=5 & 56.65 & 62.67 & 59.31 & 81.74 & 81.62 & 81.87 & 82.77 & - & 82.57  & - & - & 82.63 & 82.90 & - & 82.52   \\ 
        
        $s$=3 & 28.78 & 50.34 & 52.09 & - & - & 74.69 & - & - & 77.45  & - & - & 78.69 & - & - & 78.39  \\ 
        
        $s$=2 & - & 31.77 & 40.34 & - & 61.66 & 62.52 & 71.52 & - & 71.6  & - & - & 76.06 & - & - & 76.77 \\ \midrule

        \multirow{2.5}{*} {\emph{$\eta$=0.1}} & \multicolumn{3}{c|}{$E$=1}                      & \multicolumn{3}{c|}{$E$=5} & \multicolumn{3}{c|}{$E$=10} & \multicolumn{3}{c|}{$E$=15} & \multicolumn{3}{c}{$E$=20}  \\ \cmidrule{2-16}         
        & BABU & AVG & FN & BABU & AVG & FN & BABU & AVG & FN & BABU & AVG & FN & BABU & AVG & FN \\ \hline     

        $s$=10 & 74.58 & 72.59 & 70.30 & 83.90 & - & 83.72 & 84.05 & - & 82.71  & - & - & 82.71 & - & - & 82.35   \\ 
        
        $s$=5 & 60.35 & 61.88 & 59.01 & - & - & 80.08 & - & - & 81.79  & - & - & 81.54 & - & - & 81.77   \\ 
        
        $s$=3 & 25.06 & 28.36 & 47.54 & - & - & 65.51 & - & - & 76.06  & - & - & 76.07 & -& -  & 76.19   \\ 
        
        $s$=2 &  -  & - & 41.27 &  - & - & 58.43 & - & - & 70.86  & - & - & 73.10 & - & - & 74.90   \\ \bottomrule
     
        \end{tabular}
        } 
\label{apptab:grid_vgg}
\end{table}

\begin{table}[htp]
    \centering
    \caption{Grid Search Results for MobileNet on CIFAR-100.}
    \resizebox{\textwidth}{!}{
        \begin{tabular}{c|ccc|ccc|ccc|ccc|ccc}
        \toprule
        \multirow{2.5}{*} {\emph{$\eta$=0.1}} & \multicolumn{3}{c|}{$E$=1}                      & \multicolumn{3}{c|}{$E$=5} & \multicolumn{3}{c|}{$E$=10} & \multicolumn{3}{c|}{$E$=15} & \multicolumn{3}{c}{$E$=20}  \\ \cmidrule{2-16}         
        & BABU & AVG & FN & BABU & AVG & FN & BABU & AVG & FN & BABU & AVG & FN & BABU & AVG & FN \\ \midrule

        $s$=100 & 41.32 & 41.57 & 76.61 & \textbf{40.49} & \textbf{43.19} & 51.18 & 37.03 & 38.13 & 45.26 & 36.16 & 36.10 & 44.03 & 35.91 & 36.19 & 42.25   \\ 
        
        $s$=50 & 40.86 & 38.35 & 7.51 & \textbf{40.76} & \textbf{40.63} & 51.11 & 36.79 & 37.77 & 46.74  & 36.10 & 37.54 & 44.47 & 34.93 & 36.05 & 43.61   \\ 
        
        $s$=10 & 35.95 & 27.02 & 4.85 & \textbf{45.56} & \textbf{36.60} & 43.18 & 41.01 & 34.67 & 47.63  & 37.87 & 35.53 & 45.50 & 38.28 & 35.94 & 47.41   \\ 
        \midrule
        \multirow{2}{*} {\emph{$\eta$=0.3}} & \multicolumn{3}{c|}{$E$=1}                      & \multicolumn{3}{c|}{$E$=5} & \multicolumn{3}{c|}{$E$=10} & \multicolumn{3}{c|}{$E$=15} & \multicolumn{3}{c}{$E$=20}  \\ \cmidrule{2-16}         
        & BABU & AVG & FN & BABU & AVG & FN & BABU & AVG & FN & BABU & AVG & FN & BABU & AVG & FN \\ \midrule

        $s$=100 & 46.65 & 41.76 & 19.52 & 40.57 & 46.67 & 50.80 & 37.79 & 41.28 & 44.34  & 36.16 & 38.29 & 41.18 & 35.91 & 38.22 & 39.38   \\ 
        
        $s$=50 & 44.03 & 37.83 & 19.63 & 40.02 & 45.28 & 49.92 & 37.28 & 42.65 & 43.62  & 36.10 & 41.66 & 41.71 & 34.93 & 40.61 & 38.08   \\   
        
        $s$=10 & 37.37 & 26.66 & 16.66 & 47.07 & 33.78 & 48.89 & 44.57 & 36.42 & 46.77  & 37.87 & - & 47.95 & 38.28 & 35.12 & 48.39   \\         
        \midrule
        \multirow{2}{*} {\emph{$\eta$=0.5}} & \multicolumn{3}{c|}{$E$=1}                      & \multicolumn{3}{c|}{$E$=5} & \multicolumn{3}{c|}{$E$=10} & \multicolumn{3}{c|}{$E$=15} & \multicolumn{3}{c}{$E$=20}  \\ \cmidrule{2-16}         
        & BABU & AVG & FN & BABU & AVG & FN & BABU & AVG & FN & BABU & AVG & FN & BABU & AVG & FN \\ \midrule

        $s$=100 & 47.18 & 35.81 & 28.46 & 39.33 & - & \textbf{51.51} & 38.23 & - & 44.54  & 35.59 & - & 39.56 & 35.99 & - & 38.20   \\ 
        
        $s$=50 & 47.53 & 36.28 & 25.71 & 40.42 & - & \textbf{50.6} & 37.71 & 44.76 & 42.47  & 37.66 & - & 40.96 & 38.38 & 43.15 & 38.42   \\   
        
        $s$=10 & 38.91 & 19.51 & 21.02 & 46.48 & 11.83 & \textbf{46.87} & 45.74 & 23.02 & 44.25  & 43.61 & 24.74 & 45.05 & 42.62 & 30.68 & 43.28   \\ 
        \midrule
        \multirow{2.5}{*} {\emph{$\eta$=1.0}} & \multicolumn{3}{c|}{$E$=1}                      & \multicolumn{3}{c|}{$E$=5} & \multicolumn{3}{c|}{$E$=10} & \multicolumn{3}{c|}{$E$=15} & \multicolumn{3}{c}{$E$=20}  \\ \cmidrule{2-16}         
        & BABU & AVG & FN & BABU & AVG & FN & BABU & AVG & FN & BABU & AVG & FN & BABU & AVG & FN \\ \midrule

        $s$=100 & 46.92 & 31.15 & 36.07 & 39.61 & 34.66 & 50.95 & 37.12 & 42.06 & 42.17  & 37.83 & 46.53 & 39.64 & 37.64 & 33.12 & 39.97   \\ 
        
        $s$=50 & 45.24 & 26.64 & 32.92 & 40.11 & 28.02 & 48.46 & 38.76 & 39.47 & 40.82 & 41.13 & 42.79 & 41.07 & 39.10 & 42.41 & 35.88   \\ 
        
        $s$=10 & 38.59 & 8.39 & 26.63 & 45.14 & - & 39.64 & 44.03 & - & 38.64  & 40.51 & - & 36.05 & 39.67 & 12.26 & 37.41   \\ \bottomrule

        \end{tabular}
        }
\label{apptab:grid_mobile}
\end{table}
\newpage

\begin{table}[htp]
    \centering
    \caption{Grid Search Results for ResNet18 on CIFAR-10.} 
    \resizebox{\textwidth}{!}{
        \begin{tabular}{c|ccc|ccc|ccc|ccc}
        \toprule
        \multirow{2.5}{*} {\emph{$\eta$=0.01}} & \multicolumn{3}{c|}{$E$=5} & \multicolumn{3}{c|}{$E$=10} & \multicolumn{3}{c|}{$E$=15} & \multicolumn{3}{c}{$E$=20}  \\ \cmidrule{2-13}         
        & BABU & AVG & FN & BABU & AVG & FN & BABU & AVG & FN & BABU & AVG & FN  \\ \midrule 
        $s$=10 &  75.32  & 76.07 & 78.00 &  74.53 & 74.17 & 77.72 & \textbf{73.55} & \textbf{73.85} & 77.80  & 73.16 & 72.32 & 77.18    \\
        
        $s$=5 &  68.02  & 69.34 & 72.14 &  67.54 & 68.13 & 71.17 & \textbf{68.36} & \textbf{67.71} & 73.06  & 68.56 & 66.48 & 74.74    \\ 
        
        $s$=3 &  58.60  & 58.93 & 64.94 &  57.87 & 58.45 & 66.45 & \textbf{57.07} & \textbf{57.38} & 64.23  & 60.93 & 55.13 & 66.58    \\ 
        
        $s$=2 &  46.79  & 41.15 & 45.50 &  45.50 & 41.92 & 47.22 & \textbf{49.21} & \textbf{45.92} & 50.81  & 47.50 & 45.23 & 47.99    \\ \midrule
            
        \multirow{2}{*} {\emph{$\eta$=0.03}} & \multicolumn{3}{c|}{$E$=5} & \multicolumn{3}{c|}{$E$=10} & \multicolumn{3}{c|}{$E$=15} & \multicolumn{3}{c}{$E$=20}  \\ \cmidrule{2-13}         
        & BABU & AVG & FN & BABU & AVG & FN & BABU & AVG & FN & BABU & AVG & FN  \\ \midrule  
        $s$=10 &  78.64  & 78.77 & 80.63 &  77.76 & 77.73 & 79.86 & 76.83 & 76.57 & 79.34  & 76.02 & 76.87 & 78.78    \\
        
        $s$=5 &  70.04  & 71.50 & 75.57 & 72.01 & 71.84 & 75.77 & 71.45 & 71.72 & 75.99  & 72.64 & 71.83 & 75.94    \\ 
        
        $s$=3 &  57.53  & 61.54 & 67.55 & 56.60 & 59.58 & 66.88 & 58.06 & 60.53 & 68.63  & 60.80 & 62.07 & 68.65    \\ 
        
        $s$=2 &  47.56  & 40.02 & 52.70 & 46.26 & 46.27 & 50.25 & 50.16 & 45.60 & 48.52  & 50.60 & 46.01 & 52.37    \\ \midrule
        
        \multirow{2.5}{*} {\emph{$\eta$=0.05}} & \multicolumn{3}{c|}{$E$=5} & \multicolumn{3}{c|}{$E$=10} & \multicolumn{3}{c|}{$E$=15} & \multicolumn{3}{c}{$E$=20}  \\ \cmidrule{2-13}         
        & BABU & AVG & FN & BABU & AVG & FN & BABU & AVG & FN & BABU & AVG & FN  \\ \midrule 
        $s$=10 &  80.17  & 79.41 & 81.21 & 78.22 & 77.93 & 80.67 & 77.87 & 78.50 & 80.98  & 77.16 & 77.77 & 79.59    \\
        
        $s$=5 &  70.65  & 72.44 & 77.25 & 72.08 & 71.08 & 76.65 & 73.21 & 71.85 & 77.41  & 74.16 & 72.52 & 77.19    \\ 
        
        $s$=3 &  56.26  & 62.22 & 66.91 & 56.22 & 61.38 & 67.01 & 59.22 & 60.85 & 64.09  & 61.79 & 57.00 & 64.51    \\ 
        
        $s$=2 &  53.18  & 43.99 & 50.88 & 45.76 & 45.26 & 51.06 & 46.53 & 40.44 & 51.28  & 46.49 & 41.58 & 56.41    \\ \midrule

        \multirow{2.5}{*} {\emph{$\eta$=0.1}} & \multicolumn{3}{c|}{$E$=5} & \multicolumn{3}{c|}{$E$=10} & \multicolumn{3}{c|}{$E$=15} & \multicolumn{3}{c}{$E$=20}  \\ \cmidrule{2-13}         
        & BABU & AVG & FN & BABU & AVG & FN & BABU & AVG & FN & BABU & AVG & FN  \\ \hline  
        $s$=10 &  80.83  & 80.13  & 82.37 & 80.04 & 79.87 & 82.01 & 79.50 & 78.57 & \textbf{81.19}  & 78.64 & 78.21 & 80.37    \\
        
        $s$=5 & 72.81  & 71.56 & 77.91 & 74.37 & 69.65 & 76.89 & 74.71 & 69.66 & \textbf{75.98}  & 73.97 & 71.65 & 77.47    \\ 
        
        $s$=3 &  57.14  & 59.65 & 63.75 & 57.33 & 52.71 & 61.12 & 60.34 & 58.13 & \textbf{65.09}  & 58.46 & 54.72 & 67.08    \\ 
        
        $s$=2 & 47.35 & 50.51 & 50.58 & 42.78 & 41.36 & 51.19 & 47.63 & 44.81 & \textbf{50.12}  & 40.23 & 39.03 & 49.05    \\ \bottomrule
     
        \end{tabular}
        }
        \label{apptab:grid_resnet}
\end{table}

\begin{table}[htp]
    \centering
    \caption{Grid Search Results for Pretrained ResNet18 on CIFAR-10.}
    \resizebox{\textwidth}{!}{
        \begin{tabular}{c|ccc|ccc|ccc|ccc}
        \toprule
        \multirow{2.5}{*} {\emph{$\eta$=0.01}} & \multicolumn{3}{c|}{$E$=5} & \multicolumn{3}{c|}{$E$=10} & \multicolumn{3}{c|}{$E$=15} & \multicolumn{3}{c}{$E$=20}  \\ \cmidrule{2-13}         
        & BABU & AVG & FN & BABU & AVG & FN & BABU & AVG & FN & BABU & AVG & FN  \\ \hline  
        $s$=10 &  83.85  & 83.80 & 84.65 & 84.06 & 84.29 & 84.20 & \textbf{84.02} & \textbf{83.96} & 84.43  & 84.43 & 84.33 & 84.81    \\
        
        $s$=5 &  68.83  & 73.31 & 74.90 & 69.73 & 69.24 & 74.72 & \textbf{69.44} & \textbf{70.77} & 75.26  & 68.58 & 70.89 & 76.75    \\ 
        
        $s$=3 & 54.21 & 58.52 & 63.87 & 51.85 & 48.52 & 61.76 & \textbf{48.96} & \textbf{57.85} & 60.72  & 47.84 & 51.11 & 62.06    \\ 
        
        $s$=2 & 48.21  & 33.84 & 47.74 & 43.02 & 35.91 & 47.30 & \textbf{49.00} & \textbf{39.04} & 50.82  & 44.91 & 36.95 & 49.98    \\ \midrule
            
        \multirow{2.5}{*} {\emph{$\eta$=0.03}} & \multicolumn{3}{c|}{$E$=5} & \multicolumn{3}{c|}{$E$=10} & \multicolumn{3}{c|}{$E$=15} & \multicolumn{3}{c}{$E$=20}  \\ \cmidrule{2-13}         
        & BABU & AVG & FN & BABU & AVG & FN & BABU & AVG & FN & BABU & AVG & FN  \\ \hline  
        $s$=10 &  84.86  & 82.73 & 85.28 & 85.02 & 82.00 & 85.07 & 84.85 & 82.71 & 85.10  & 84.82 & 83.04 & 85.27    \\
        
        $s$=5 & 67.27  & 60.13 & 72.24 & 68.64 & 59.54 & 70.55 & 67.81 & 65.43 & 73.14  & 68.23 & 68.91 & 73.27    \\ 
        
        $s$=3 & 53.48 & 47.72 & 54.61 & 53.74 & 50.32 & 56.27 & 54.18 & 53.78 & 59.92  & 56.35 & 53.82 & 60.28    \\ 
        
        $s$=2 & 43.10  & 36.69 & 46.89 &  45.80 & 32.61 & 49.99 & 51.62 & 39.08 & 45.50 & 55.78 & 41.35 & 51.88    \\ \midrule
        
        \multirow{2.5}{*} {\emph{$\eta$=0.05}} & \multicolumn{3}{c|}{$E$=5} & \multicolumn{3}{c|}{$E$=10} & \multicolumn{3}{c|}{$E$=15} & \multicolumn{3}{c}{$E$=20}  \\ \cmidrule{2-13}         
        & BABU & AVG & FN & BABU & AVG & FN & BABU & AVG & FN & BABU & AVG & FN  \\ \hline  
        $s$=10 &  86.06  & 71.37 & 85.84 & 85.71 & 75.94 & 85.52 & 85.31 & 76.50 & 84.73  & 85.41 & 77.77 & 85.11    \\
        
        $s$=5 &  69.47  & 65.13 & 69.89 & 69.72 & 60.01 & 71.02 & 69.49 & 62.47 & 73.75  & 70.17 & 68.17 & 73.24    \\ 
        
        $s$=3 & 57.91 & 52.70 & 53.10 & 55.63 & 49.75 & 54.86 & 62.72 & 54.51 & 63.03  & 65.22 & 52.00 & 65.71    \\ 
        
        $s$=2 & 44.49 & 42.07 & 45.55 & 43.65 & 43.69 & 45.46 & 48.36 & 37.11 & 52.96  & 50.87 & 41.37 & 51.87    \\ \midrule

        \multirow{2.5}{*} {\emph{$\eta$=0.1}} & \multicolumn{3}{c|}{$E$=5} & \multicolumn{3}{c|}{$E$=10} & \multicolumn{3}{c|}{$E$=15} & \multicolumn{3}{c}{$E$=20}  \\ \cmidrule{2-13}         
        & BABU & AVG & FN & BABU & AVG & FN & BABU & AVG & FN & BABU & AVG & FN  \\ \hline  
        $s$=10 &  85.89  & 72.78 & 86.02 & 85.67 & 74.95 & 85.65 & 84.98 & 78.00 & \textbf{85.45}  & 85.25 & 78.73 & 85.20    \\
        
        $s$=5 &  68.32  & 65.71 & 74.01 & 70.17 & 64.67 & 73.20 & 75.10 & 67.00 & \textbf{73.43}  & 73.03 & 71.16 & 77.12    \\ 
        
        $s$=3 & 62.57 & 46.52 & 66.26 & 64.15 & 59.18 & 74.36 & 63.56 & 56.59 & \textbf{73.85}  & 66.52 & 59.25 & 73.86    \\ 
        
        $s$=2 & 45.93 & 38.19 & 41.82 & 45.26 & 38.53 & 50.13 & 52.22 & 38.06 & \textbf{50.94}  & 55.18 & 47.67 & 53.42    \\ \bottomrule
     
        \end{tabular}
        }
        \label{apptab:grid_resnet_pretrained}
\end{table}
\clearpage
\subsection{SphereFed}
We conduct grid searches for SphereFed (CE) and SphereFed (MSE)~\citep{dong2022spherefed} for MobileNet on CIFAR-100. In Table ~\ref{apptab:grid_sphere}, we conduct a $\eta$ search within the range $\{0.1, 0.3, 0.5, 1.0\}$ for both SphereFed (CE) and SphereFed (MSE). However, SphereFed (MSE) exhibit suboptimal performance within this range. Consequently, in Table ~\ref{apptab:grid_sphere_additional}, we extend the $\eta$ search exclusively for SphereFed (MSE) to include values $\{1.5, 3.0, 4.5, 5.0\}$. In summary, the determined optimal hyperparameter $\eta$ values are 0.03 for SphereFed (CE) and 4.5 for SphereFed (MSE).

\begin{table}[htp]
    \centering
        \caption{Grid Search Results for SphereFed with MobileNet on CIFAR-100.}
        \resizebox{0.6\textwidth}{!}{
        \begin{tabular}{c|cc|cc|cc}
        \toprule
        \multirow{2.5}{*} {\emph{$\eta$=0.1}} & \multicolumn{2}{c|}{$E$=1}                      & \multicolumn{2}{c|}{$E$=5} & \multicolumn{2}{c}{$E$=10} \\ \cmidrule{2-7}         
        & CE & MSE & CE & MSE & CE & MSE  \\  
        \midrule
        $s$=100 & 12.96 & 1.37 & 44.92 & 2.16 & 39.50 & 4.25 \\ 
        $s$=50 & 12.21 & 1.30 & 43.38 & 1.73 & 38.65 & 3.84 \\ 
        $s$=10 & 6.67 & 1.46 & 33.32 & 2.34 & 30.02 & 3.81     \\ \midrule        
        \multirow{2.5}{*} {\emph{$\eta$=0.3}} & \multicolumn{2}{c|}{$E$=1}                      & \multicolumn{2}{c|}{$E$=5} & \multicolumn{2}{c}{$E$=10} \\ \cmidrule{2-7}       
        & CE & MSE & CE &  MSE & CE & MSE  \\ \hline 
        $s$=100 & 24.62 & 1.39 & \textbf{44.96} & 5.25 & 40.47 & 19.41     \\ 
        $s$=50 & 23.56 & 1.44 & \textbf{43.68} & 5.89 & 38.21 & 19.65     \\   
        $s$=10 & 14.91 & 1.40 & \textbf{40.39} & 4.46 & 33.53 & 15.69     \\ \midrule        
        \multirow{2.5}{*} {\emph{$\eta$=0.5}} & \multicolumn{2}{c|}{$E$=1}                      & \multicolumn{2}{c|}{$E$=5} & \multicolumn{2}{c}{$E$=10} \\ \cmidrule{2-7}       
        & CE & MSE & CE &  MSE & CE & MSE  \\ \hline         
        $s$=100 & 32.12 & 1.46 & 45.47 & 12.30 & 40.11 & 36.38     \\    
        $s$=50 & 28.45 & 1.43 & 44.97 & 11.76 & 39.49 & 34.07     \\   
        $s$=10 & 18.70 & 1.61 & 37.02 & 8.32 & 33.83 & 24.65     \\ \midrule
        \multirow{2}{*} {\emph{$\eta$=1.0}} & \multicolumn{2}{c|}{$E$=1}                      & \multicolumn{2}{c|}{$E$=5} & \multicolumn{2}{c}{$E$=10} \\ \cmidrule{2-7}       
        & CE & MSE & CE &  MSE & CE & MSE  \\ \midrule
        $s$=100 & 38.98 & 2.33 & 46.45 & 29.94 & 41.58 & 40.02 \\
        $s$=50 & 35.69 & 2.27 & 43.95 & 29.39 & 41.20 & 40.87     \\ 
        $s$=10 & 26.40 & 1.70 & 36.72 & 23.38 & 35.73 & 32.64     \\ \bottomrule
        \end{tabular}
        }
        \label{apptab:grid_sphere}
\end{table}

\begin{table}[htp]
    \centering
    \caption{Additional Grid Search Results for SphereFed (MSE) with MobileNet on CIFAR-100.}
    \resizebox{0.3\textwidth}{!}{
        \begin{tabular}{cccc}
        \toprule
        \multirow{2.5}{*} {\emph{$\eta$=1.5}} & {$E$=1}                      & {$E$=5} & {$E$=10} \\ \cmidrule{2-4}         
        &  MSE &   MSE &  MSE  \\ \midrule

        $s$=100 & 2.84  & 44.40 & 40.32     \\ 
        
        $s$=50 & 2.81  & 41.91 & 42.18     \\  
        
        $s$=10 & 2.69  & 33.68 & 35.66     \\ \midrule
        \multirow{2.5}{*} {\emph{$\eta$=3.0}} & {$E$=1}                      & {$E$=5} & {$E$=10} \\ \cmidrule{2-4}         
        &  MSE &   MSE &  MSE  \\ \midrule

        $s$=100 & 6.37  & 47.44 & 42.15     \\ 
        
        $s$=50 & 5.05  & 45.56 & 42.61     \\  
        
        $s$=10 & 4.87  & 40.84 & 34.68     \\ \midrule

        \multirow{2.5}{*} {\emph{$\eta$=4.5}} & {$E$=1}                      & {$E$=5} & {$E$=10} \\ \cmidrule{2-4}         
        &  MSE &   MSE &  MSE  \\ \midrule        
         
        $s$=100 & 9.43  & \textbf{49.51} & 43.19     \\ 
        
        $s$=50 & 8.19  & \textbf{48.05} & 42.42     \\ 
        
        $s$=10 & 7.50  & \textbf{43.42} & 37.93     \\ \midrule

        \multirow{2.5}{*} {\emph{$\eta$=5.0}} & {$E$=1}                      & {$E$=5} & {$E$=10} \\ \cmidrule{2-4}         
        &  MSE &   MSE &  MSE  \\ \midrule         
        $s$=100 & 10.58  & 49.29 & 36.07     \\ 
        
        $s$=50 & 10.95  & 48.08 & 38.49     \\  
        
        $s$=10 & 8.26  & 43.26 & 21.73     \\ \bottomrule
        \end{tabular}
        }
        \label{apptab:grid_sphere_additional}
\end{table}
\clearpage
\section{Additional Experiments}
\label{app:add_exp}
\subsection{Weight Norm Does Not Matter on FedFN}
\begin{figure}[h!]
    \centering
    \makebox[\textwidth]{
    \includegraphics[width=0.6\textwidth]{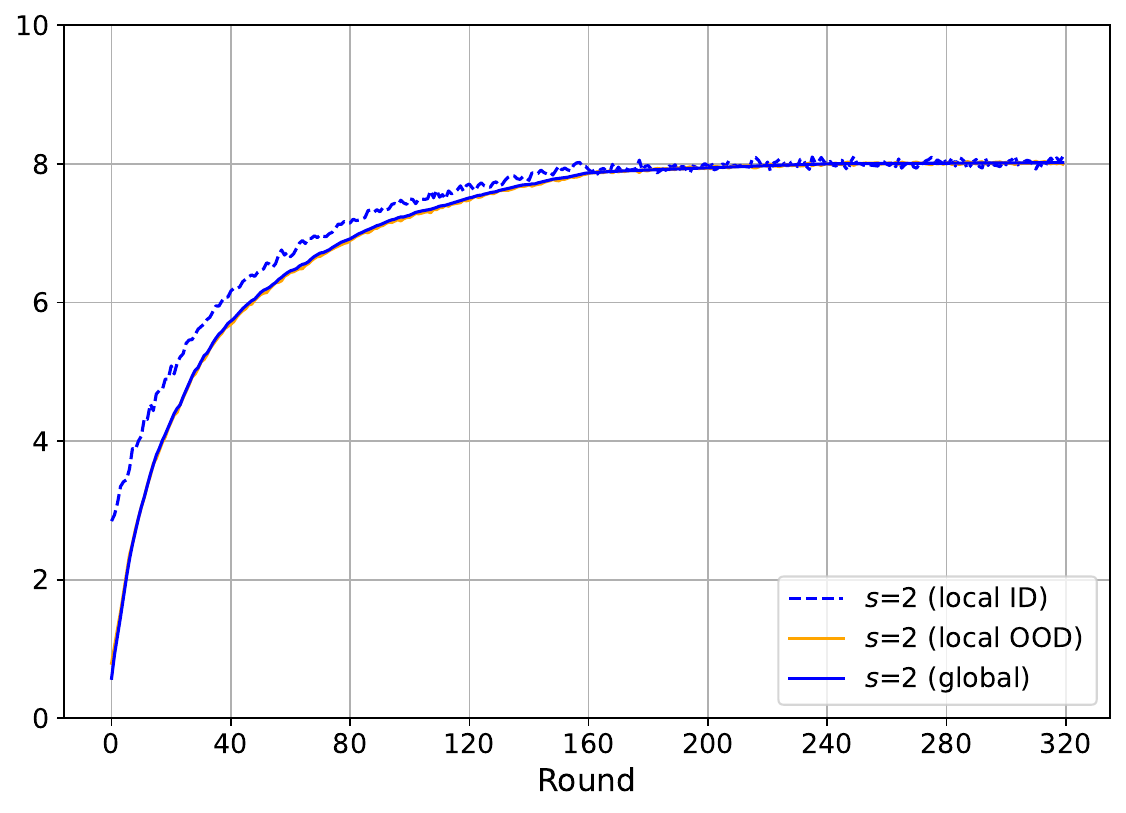}
    }
    \vspace{-0.3 in}
    \caption{Discrepancy in Weight Norms between Local and Global Models in the $s$=2 Setting.}
    \label{appfig:weight_norm_bias_fn}
\end{figure}

We investigate the impact of increased data heterogeneity on weight norm disparity in the case of FedFN. Focusing on the $s$=2 setting, representing the most heterogeneous scenario, as shown in Figure~\ref{appfig:weight_norm_bias_fn}, we explore the evolution of weight norms between global and local models. Initially, during early training stages, a noticeable weight norm bias emerges within local models, favoring seen (ID) classes over unseen (OOD) classes. However, this bias progressively diminishes as the learning rate decreases, eventually aligning local models with the weight norm mean of the global model.

\begin{figure}[h!]
    \centering
    \vspace{-0.15 in}
    \makebox[\textwidth]{
    \includegraphics[width=0.8\textwidth]{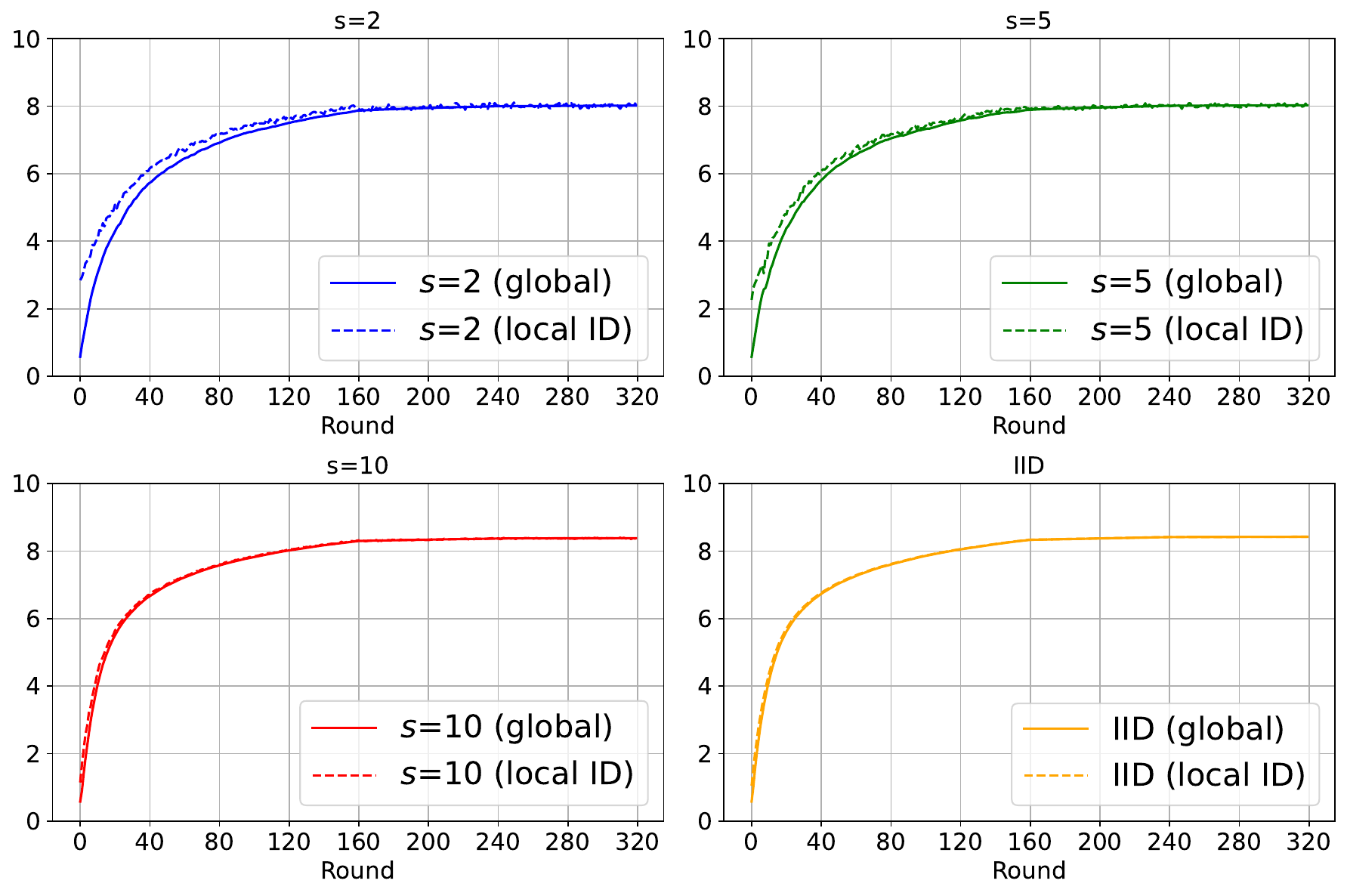}
    }
    \vspace{-0.2 in}
    \caption{Discrepancy in Weight Norms between Local and Global Models in Various Heterogeneity Settings.}
    \label{appfig:weight_norm_disparity_fn}
\end{figure}

Additionally, we analyze variations in weight norm means between local models on the ID classes and the global model on the total classes under varying data heterogeneity. Specifically, we consider $s \in \{2, 5, 10\}$ and IID (Exactly class balanced data distribution across clients) on the CIFAR-10 dataset. As illustrated in Figure~\ref{appfig:weight_norm_disparity_fn}, during initial training phases, the norm disparity between the global model and local models increases with data heterogeneity (i.e., IID $\rightarrow$ $s$=10 $\rightarrow$ $s$=5 $\rightarrow$ $s$=2). However, this disparity gradually diminishes across all settings as the learning rate decreases.

\newpage
\subsection{Additional GFL Results}

In the main paragraph of Section~\ref{sec:exp}, we report results exclusively in a balanced environment, specifically in the sharding setting. In Table~\ref{tab:gfl_acc_all}, results are presented, encompassing an unbalanced environment, specifically in the LDA setting. The reported outcomes stem from implementations with a consistent and identical seed, deviating from the context of the paragraph. In contrast to the main paragraph, we observe that in the MobileNet on CIFAR-100 environment with the $s=10$ setting, all baseline algorithms exhibit notably lower performance for the seed used in Table~\ref{tab:gfl_acc_all}. However, when applying the BABU or FN module, the performance difference is less pronounced, showing a more stable outcome compared to the main paragraph. Furthermore, consistently across all settings, applying the FN module to the baseline consistently demonstrates superior performance.

\begin{table}[h!]
    \centering
    \vspace{-0.15 in}
    \caption{FL Accuracy Comparison for Baseline, +BABU, and +FN.}
    \small
    \resizebox{1.0\textwidth}{!}{
    \begin{tabular}{cc|cccc|cccc|ccc|cccc}
      \toprule 
      \multirow{2.5}{*}{Algorithm} & \multirow{2.5}{*}{Module} & \multicolumn{8}{c|}{VGG11 on CIFAR-10} & \multicolumn{7}{c}{MobileNet on CIFAR-100} \\ \cmidrule{3-17}
      & & $s$=2 & $s$=3 & $s$=5 & $s$=10 & $\alpha$=0.1 & $\alpha$=0.3 & $\alpha$=0.5 & $\alpha$=1.0 & $s$=10 & $s$=50 & $s$=100 & $\alpha$=0.1 & $\alpha$=0.3 & $\alpha$=0.5 & $\alpha$=1.0 \\ \midrule
      & Baseline  & 73.83 & 78.99 & 81.40 & 82.79 & 73.16 & 80.77 & 81.63 & 82.45 & 26.64 & 40.18 & 41.50 & 42.66 & 42.68 & 43.82 & 41.57  \\ 
      FedAVG  & + BABU & 74.31 & 78.88 & 81.17 & 82.37  & 72.42 & 80.59 & 80.94 & 82.93 &  43.96 & 39.94 & 40.37 & 42.99 & 39.61 & 39.53 & 39.58  \\ 
      & \textbf{+ FN} & \textbf{77.23} & \textbf{81.46} & \textbf{82.80} & \textbf{84.16}  & \textbf{75.58} & \textbf{81.85} & \textbf{83.06} & \textbf{83.65} & \textbf{45.16} & \textbf{48.83} & \textbf{49.54} & 47.87 & 47.66 & 47.86 & 47.80  \\  \midrule
      
      & Baseline & 77.07 & (\textit{Failed}) & \textbf{84.50} & 85.28 & (\textit{Failed}) & 83.24 & 83.54 & 85.21 & 34.64 & 42.56 & 43.67 & \textbf{45.18} & 45.62 & 44.42 & 43.16  \\
      Scaffold & + BABU & 76.89 & 82.24 & 84.26 & 85.02 & (\textit{Failed}) & 82.95 & 84.12 & 85.06 & 46.80 & 42.73 & 44.50 & 43.28 & 43.76 & 44.37 & 44.20 \\ 
       & \textbf{+ FN} & \textbf{79.05} & \textbf{82.83} & \textbf{84.80} & \textbf{85.83} & \textbf{74.53} & \textbf{83.65} & \textbf{84.79} & \textbf{85.42} & \textbf{50.17} & \textbf{48.74} & \textbf{51.04} & 45.10 & \textbf{49.71} & \textbf{50.42} & \textbf{52.13}  \\  \midrule

       
      & Baseline & 73.92 & 78.86 & 81.47 & 82.61 & 73.39 & 80.94 & 81.16 & 82.64 & 26.57 & 39.93 & 42.11 & 42.71 & 42.77 & 44.54 & 41.55 \\
      FedEXP & +BABU & 74.48 & 78.91 & 81.16 & 82.17 & 71.61 & 80.50 & 81.24 & 82.72 & 45.44 & 41.49 & 40.24 & 42.73 & 40.09 & 39.54 & 40.66  \\ 
      & \textbf{+ FN} & \textbf{77.17} & \textbf{80.92} & \textbf{82.79} & \textbf{83.57} & \textbf{75.03} & \textbf{82.29} & \textbf{83.12} & \textbf{83.36} & \textbf{45.37} & \textbf{48.45} & \textbf{49.80} & \textbf{47.73} & \textbf{47.16} & \textbf{48.32} & \textbf{48.01} \\  \bottomrule 
    \end{tabular}
    }
    \label{tab:gfl_acc_all}
\end{table}
\subsection{Personalized Federated Learning (PFL) Results}\label{subsec:pfl_result}

We present FedFN-FT, a fine-tuned algorithm for PFL, inspired by prior work~\citep{oh2021fedbabu, dong2022spherefed}, utilizing local data.  We compare FedFN-FT with existing PFL methods, including simple local models, 1-step approaches like FedPer~\cite{arivazhagan2019federated}, Per-FedAVG~\cite{fallah2020personalized}, and FedRep~\cite{collins2021exploiting}, as well as 2-step methods such as FedAVG-FT, FedBABU-FT~\cite{oh2021fedbabu}, SphereFed-FT (CE, MSE)\cite{dong2022spherefed}. Table~\ref{apptab:pfl_acc} provides detailed personalized accuracy results. The entries form of X±Y, representing the mean and standard deviation of personalized accuracies across all clients for PFL algorithms. Entries without standard deviation indicate performance on $D_{test}$, derived from the global model after the initial step of 2-step methods.

Regarding the SphereFed method, it constructs each element of the logit vector based on the cosine similarity between feature vectors and classifiers. Similar to FedFN, SphereFed necessitates rescaling the learning rate for classifier weights, leading to the need for separate learning rate tuning.  To address this requirement,  we conduct an extensive grid search to determine the appropriate initial learning rate, denoted as $\eta$. Further details about the grid search are provided in Section~\ref{app:grid search}. 

For SphereFed (CE), SphereFed (MSE), and FedFN, we initialize the learning rates with $\eta$ values of 0.3, 4.5, and 0.5, respectively. The 2-step fine-tuning methods undergo a total of 5 local epochs, during which learning rate was carried out through grid search within the range of $\{\eta, 0.1\times\eta, 0.01\times\eta\}$. The comprehensive results of the grid search for SphereFed and FN, including learning rate adjustments, are confirmed to be available in Appendix~\ref{app:finetuneing_lr_search}. The resulting tunned learning rates are documented in Table~\ref{apptab:pfl_acc}. In summary, 2-step methods consistently outperform 1-step methods in terms of PFL performance. Among the 2-step methods, FedFN-FT consistently exhibits superior performance.

\begin{table}[htp]
    \centering
    \caption{PFL Accuracy Comparison for MobileNet on CIFAR-100.}
    \small
    \begin{tabular}{l|ccc}
    \toprule
    Algorithm & $s$=10 & $s$=50 & $s$=100 \\ \midrule
    Local only                              & 58.64{\tiny $\pm$7.21}  & 25.38{\tiny $\pm$4.12} & 18.52{\tiny $\pm$3.15} \\ \midrule
    FedPer (2019)                           & 70.92{\tiny $\pm$6.93}  & 33.73{\tiny $\pm$4.68} & 22.77{\tiny $\pm$4.30} \\
    Per-FedAVG (2020)                       & 32.57{\tiny $\pm$11.02} & 43.09{\tiny $\pm$7.36} & 45.00{\tiny $\pm$7.05} \\
    FedRep (2021)                           & 62.69{\tiny $\pm$7.26}  & 34.66{\tiny $\pm$5.34} & 26.53{\tiny $\pm$4.52} \\ \midrule
    FedAVG (2017)                           & 36.30 & 41.97   & 42.51 \\ 
    FedAVG-FT (0.1$\times\eta$)             & 77.39{\tiny $\pm$6.40}  & 50.57{\tiny $\pm$5.31} & 46.95{\tiny $\pm$4.90} \\
    FedBABU (2022)                          & 45.73 & 39.57   & 40.70  \\
    FedBABU-FT (0.01$\times\eta$)           & 79.76{\tiny $\pm$6.05}  & 50.93{\tiny $\pm$4.69} & 46.63{\tiny $\pm$5.25} \\
    SphereFed (CE) (2022)                   & 40.39 & 43.68   & 44.96 \\ 
    SphereFed-FT (CE) (0.1$\times\eta$)     & 77.24{\tiny $\pm$6.38}  & 55.08{\tiny $\pm$5.33} & 50.17{\tiny $\pm$5.04} \\ 
    SphereFed (MSE) (2022)                  & 43.42 & 48.05   & 49.51 \\ 
    SphereFed-FT (MSE) (0.01$\times\eta$)   & \textbf{82.50}{\tiny $\pm$6.04} & 56.45{\tiny $\pm$5.56} & 53.10{\tiny $\pm$4.90}   \\ \midrule
    FedFN                                   & 46.98 & 49.47   & 50.92 \\
    \textbf{FedFN-FT} (0.1$\times\eta$)     & \textbf{82.85}{\tiny $\pm$5.83} & \textbf{60.78}{\tiny $\pm$4.99} & \textbf{55.43}{\tiny $\pm$5.24} \\
    \bottomrule
    \end{tabular}
    \label{apptab:pfl_acc}
\end{table}
\newpage

\subsubsection{Fine Tuning Learning Rate Search}\label{app:finetuneing_lr_search}
\begin{table*}[h!]
  \centering  
  \caption{Search for Finetuning Learning Rates in Two-Step Methods.}
\resizebox{0.8\textwidth}{!}{%
\begin{tabular}{c|ccc}
\toprule
Algorithm &  $s$=10 & $s$=50 & $s$=100  \\ \midrule

FedAVG (2017) & 36.30    & 41.97    & 42.51        \\ 

FedAVG-FT ($\eta$) & 67.23 $\pm$ 6.32 & 35.98 $\pm$ 5.56 & 38.25 $\pm$ 5.29    \\ 

\textbf{FedAVG-FT} (0.1$\times\eta$) & \textbf{77.39 $\pm$ 6.40} & \textbf{50.57 $\pm$ 5.31} & \textbf{46.95 $\pm$ 4.90} \\

FedAVG-FT (0.01$\times\eta$) & 70.96 $\pm$ 6.81   & 44.78 $\pm$ 4.88   & 44.00 $\pm$ 4.93       \\ 
\midrule

FedBABU (2022) & 45.73 & 39.57  & 40.7  \\ 

FedBABU-FT ($\eta$) & 13.32 $\pm$ 4.51   & 4.72 $\pm$ 1.85   & 2.85 $\pm$ 1.27       \\ 

FedBABU-FT (0.1$\times\eta$) & 77.88 $\pm$ 6.59 & 51.75 $\pm$ 4.84 & 46.02 $\pm$ 5.24 \\ 

\textbf{FedBABU-FT} (0.01$\times\eta$) & \textbf{79.76 $\pm$ 6.05} & \textbf{50.93 $\pm$ 4.69} & \textbf{46.63 $\pm$ 5.25} \\ \midrule

SphereFed (CE) (2022) & 40.39  & 43.68  & 44.96  \\ 

SphereFed-FT (CE) ($\eta$) & 53.79 $\pm$ 9.47 & 32.11 $\pm$ 6.89 & 27.81 $\pm$ 4.98 \\ 

\textbf{SphereFed-FT (CE)} (0.1$\times\eta$) & \textbf{77.24 $\pm$ 6.38}   & \textbf{55.08 $\pm$ 5.33} & \textbf{50.17 $\pm$ 5.04} \\ 

SphereFed-FT (CE) (0.01$\times\eta$) & 77.18 $\pm$ 6.17 & 49.99 $\pm$ 4.79   & 47.45 $\pm$ 4.70       \\

\midrule

SphereFed (MSE) (2022) & 43.42  & 48.05  & 49.51  \\ 

SphereFed-FT (MSE) ($\eta$) & 57.34 $\pm$ 7.53 & 35.43 $\pm$ 5.32 & 27.95 $\pm$ 5.03 \\ 

SphereFed-FT (MSE) (0.1$\times\eta$) & 79.73 $\pm$ 6.63 & 56.80 $\pm$ 5.29 & 51.06 $\pm$ 5.59 \\ 

\textbf{SphereFed-FT (MSE)} (0.01$\times\eta$) & \textbf{82.50 $\pm$ 6.04} & \textbf{56.45 $\pm$ 5.56} & \textbf{53.10 $\pm$ 4.90} \\ 
\midrule
FedFN  & 46.98 & 49.47 & 50.92  \\

FedFN-FT ($\eta$) & 71.88 $\pm$ 6.73 & 43.80 $\pm$ 4.54 & 41.23 $\pm$ 5.46       \\ 

\textbf{FedFN-FT} (0.1$\times\eta$) & \textbf{82.85 $\pm$ 5.83}   & \textbf{60.78 $\pm$ 4.99} & \textbf{55.43 $\pm$ 5.24}    \\

FedFN-FT (0.01$\times\eta$) & 82.15 $\pm$ 5.74 & 54.24 $\pm$ 5.06 & 51.92 $\pm$ 5.06    \\ 
\bottomrule
   
        \end{tabular}%
    }
\label{tab:2_step_fine_tune_lr_search}
\end{table*}

\newpage
\subsection{Logit Should Be Non-Restricted}
SphereFed~\cite{dong2022spherefed} modifies i-th index of the logit vector of an input $x$, represented as $\tilde{z}_{i}(x;\theta)=\tilde{\theta}_{cls, i}\frac{f(x;\theta_{ext})}{||f(x;\theta_{ext})||_{2}}$. It maintains the classifier $\Tilde{\theta}_{cls}$ in a frozen state, ensuring that the norms of $\Tilde{\theta}_{cls, i}$ are orthonormal to each other. As a result, $\tilde{z}_{i}(x;\theta)$ becomes the cosine similarity between $f(x;\theta_{ext})$ and $\tilde{\theta}_{cls, i}$, yielding values restricted to the range [-1,1]. Following this modification, the logit margin is constrained to a maximum value of 2.

As seen in Table~\ref{apptab:pfl_acc}, despite utilizing feature normalization, SphereFed (CE) exhibits inferior performance compared to even FedBABU. To address this limitation, we propose a modification to the SphereFed (CE) logit vector. We transform it to $\tilde{z}^{\tau}_{i}(x;\theta)=\tau\,\tilde{\theta}_{cls, i}\frac{f(x;\theta_{ext})}{||f(x;\theta_{ext})||_{2}}$, which yields values in the range of [-$\tau$, $\tau$], providing less constrained outputs. We apply this approach with different values of $\tau$, specifically $\{10, 15, 20, 25, 30\}$, in the $s$=10 setting. We compared the results of this modified SphereFed (CE) with those of SphereFed (MSE), FedBABU, and FedFN, and the outcomes are presented in Table~\ref{apptab:modified_spherefed}.
Increasing $\tau$ up to 15 results in improvements in SphereFed(CE), although not as significant as compared to FedFN. However, it shows enhancements over FedBABU and SphereFed(MSE) at 15. Consequently, this indicates that creating the logit vector through feature normalization with relaxed constraints on the elements of the logit is recommended.
 
\begin{table}[h!]
\centering
\caption{Comparison of Modified SphereFed (CE) with SphereFed (MSE), FedBABU, and FedFN on $s$=10 Setting of CIFAR-100.}
\resizebox{0.4\textwidth}{!}{
\begin{tabular}{c|c}
\toprule
Algorithm & Accuracy \\ \midrule
SphereFed (CE), $\tau$=1 & 40.39 \\   
SphereFed (CE), $\tau$=10  & 42.84 \\ 
SphereFed (CE), $\tau$=15  & \textbf{45.78} \\ 
SphereFed (CE), $\tau$=20  & 44.95 \\ 
SphereFed (CE), $\tau$=25  & 44.46 \\ 
SphereFed (CE), $\tau$=30  & 39.62 \\  \midrule
SphereFed (MSE)  & 43.42 \\\midrule
FedBABU  & 45.73 \\ \midrule
FedFN  & \textbf{46.98} \\
\bottomrule
\end{tabular}
}
\label{apptab:modified_spherefed}
\end{table}

\newpage
\subsection{Reproduced Result from SphereFed}

We present the experimental results for SphereFed~\cite{dong2022spherefed} trained with LDA settings ($\alpha\in\{0.1, 0.5\}$) on CIFAR-100. To reproduce the experiments presented in the original paper, we deviated from our previous experimental settings. Specifically, for the MobileNetV2 model architecture, we constructed the layers exactly as described in Table 7 of \citep{dong2022spherefed}. Regarding the training setup, each case is trained for 500 rounds using cosine annealing, following the guideline of original paper. We follow the instructions of original paper for all other hyperparameters as well. It should be noted that we did not employ the FFC algorithm in any of the experiments, including those using FedAVG and FedFN.

Table~\ref{apptab:spherefed_reproduce} presents the new results we obtain and compares them with the original outcomes of SphereFed, encompassing FedAVG, FedFN and centralized learning. If certain algorithms are not indicated at the original results, we represent them with a dash (``-"). For the algorithms implemented as described in the original paper, we provide specific details like the actual learning rate $\eta$. Furthermore, for each FL algorithm, we present reproduced results across a specified range of initial learning rates $\eta\in\{0.1, 0.3, 0.5, 1.0, 1.5, 3.0, 4.5, 5.0\}$. In the case of centralized learning, results are specifically provided for $\eta=0.1$.

Following this, we conclude that the results of the original paper could not be reproduced. The reported performance of FL algorithms in the actual original paper (71.85, 68.78) appears surprisingly higher than the reproduced results in centralized learning (68.27). Moreover, implementing the algorithms with the exact settings reported in the original results consistently leads to lower performance (71.85 vs 18.01, 68.78 vs 37.76). Even when implemented in accordance with the specifications of the original paper ($\eta$=0.5), SphereFed (MSE) demonstrated significantly poor performance at 18.01. Subsequently, despite a thorough investigation through grid search, the best-performing configuration obtained is 52.69, still falling below the reported performance. Furthermore, when comparing the performance at the optimal learning rate for each FL algorithm, we consistently observe that FedFN outperforms the baselines.

\begin{table}[h!]
    \centering
    \caption{Reproduced Results for SphereFed, FedAVG, and FedFN under the Same Settings, Utilizing MobileNet (as Described in \citep{dong2022spherefed}) on CIFAR-100.}

    \resizebox{\textwidth}{!}{
        \begin{tabular}{c|cccccccc|c}
        \toprule
        \emph{SphereFed (MSE)} & \emph{$\eta$=0.1} & \emph{$\eta$=0.3} & \emph{$\eta$=0.5} & \emph{$\eta$=1.0} & \emph{$\eta$=1.5} & \emph{$\eta$=3.0} & \emph{$\eta$=4.5} & \emph{$\eta$=5.0} & Original Result (\emph{$\eta$=0.5})   \\ \cmidrule{1-10}         
        $\alpha$=0.5 & 2.77 & 9.73 & 18.01 & 43.26 & 52.19 & \textbf{52.69} & 48.20 & 40.87 & 71.85 \\
        $\alpha$=0.1 & 2.88 & 9.65 & 20.19 & 41.48 & 45.41 & \textbf{46.34} & 43.62 & 40.56 & - \\ \midrule
        \emph{SphereFed (CE)} & \emph{$\eta$=0.1} & \emph{$\eta$=0.3} & \emph{$\eta$=0.5} & \emph{$\eta$=1.0} & \emph{$\eta$=1.5} & \emph{$\eta$=3.0} & \emph{$\eta$=4.5} & \emph{$\eta$=5.0} & Original Result   \\ \cmidrule{1-10}         
        $\alpha$=0.5 & 37.57 & 49.39 & 48.18 & 52.51 & \textbf{52.99} & 51.09 & 44.87 & 44.23 & - \\
        $\alpha$=0.1 & 22.37 & 40.58 & 42.85 & 40.92 & \textbf{47.72} & 42.11 & 35.30 & 33.83 & - \\ \midrule
        \emph{FedAVG} & \emph{$\eta$=0.1} & \emph{$\eta$=0.3} & \emph{$\eta$=0.5} & \emph{$\eta$=1.0} & \emph{$\eta$=1.5} & \emph{$\eta$=3.0} & \emph{$\eta$=4.5} & \emph{$\eta$=5.0} & Original Result (\emph{$\eta$=0.1})   \\ \cmidrule{1-10}         
        $\alpha$=0.5 & 37.76 & \textbf{38.82} & 23.76 & 1.04 & 1.02 & 1.03 & 1.25 & 1.01 & 68.78 \\
        $\alpha$=0.1 & 38.58 & \textbf{40.47} & 27.35 & 1.22 & 1.04 & 1.02 & 1.09 & 1.17 & - \\ \midrule
        \emph{FedFN} & \emph{$\eta$=0.1} & \emph{$\eta$=0.3} & \emph{$\eta$=0.5} & \emph{$\eta$=1.0} & \emph{$\eta$=1.5} & \emph{$\eta$=3.0} & \emph{$\eta$=4.5} & \emph{$\eta$=5.0} & Original Result   \\ \cmidrule{1-10}         
        $\alpha$=0.5 & 55.00 & \textbf{53.38} & 48.55 & 49.89 & 45.79 & 42.45 & 35.43 & 34.82 & - \\
        $\alpha$=0.1 & 46.38 & \textbf{49.16} & 46.61 & 41.69 & 42.03 & 38.80 & 30.92 & 2.89 & - \\ \midrule
        \emph{Centralized Learning} &   \multicolumn{8}{c|}{\textbf{68.27} (\emph{$\eta$=0.1})} & - \\ \bottomrule

        \end{tabular}
        }
        \label{apptab:spherefed_reproduce}
\end{table}

\newpage
\subsection{FedFN vs FedFR}

We compare the performance of FedFN with Federated Averaging with Feature Norm Regularization (FedFR) introduced in Eq.~\eqref{eqn:fedfr} in the main paragraph. Table~\ref{apptab:cifar10_fedfr} and Table~\ref{apptab:cifar100_fedfr} present the performance on CIFAR-10 and CIFAR-100 with $s=10$ setting, respectively.

Table~\ref{apptab:cifar10_fedfr} reports results of FedFR on CIFAR-10, referring to the optimal hyperparameter $\mu=0.005$ from Figure~\ref{fig:fedfr-result} in the main paragraph. FedFR exhibits slightly lower performance compared to FedFN but demonstrates superiority over FedAVG and FedBABU. In the $s=10$ setting of CIFAR-100, FedFR shows comparable or superior performance to FedAVG. However, FedFR performs worse than both FedBABU and FedFN across all hyperparameter candidates. In contrast to FedFR, FedFN consistently demonstrates superior performance across all settings.

\begin{table}[htp]
    \centering
    \vspace{-0.15 in}
    \caption{Accuracy Comparison on CIFAR-10.}
    \small
    \begin{tabular}{c|cccc}
    \toprule
    \multirow{2.5}{*}{Algorithm} & \multicolumn{4}{c}{VGG11 on CIFAR-10}\\ \cmidrule{2-5} 
                                 & $s$=2 & $s$=3 & $s$=5 & $s$=10 \\ \midrule
    FedAVG   & 74.24 & 77.29 & 81.08 & 81.97  \\ 
    FedBABU   & 75.05 & 77.73 & 81.04 & 82.16  \\     
    FedFR  & 76.14 & 77.89 & 81.61 & 82.18 \\ 
    FedFN    & \textbf{77.77} & \textbf{78.93} & \textbf{82.43} & \textbf{83.80} \\ \bottomrule                            
    \end{tabular}    \label{apptab:cifar10_fedfr}
\end{table}

\begin{table}[h!]
    \centering
    \caption{Accuracy Comparison on the $s=10$ Setting of MobileNet on CIFAR-100.}
    \resizebox{\textwidth}{!}{
        \begin{tabular}{c|cccccccc}
        \toprule
        \multirow{2.5}{*}{FedFR} & \emph{$\mu$=0.5} & \emph{$\mu$=0.1} &\emph{$\mu$=0.05} & \emph{$\mu$=0.01} & \emph{$\mu$=0.005} & \emph{$\mu$=0.001} & \emph{$\mu$=0.0005} & \emph{$\mu$=0.0001}\\ \cmidrule{2-9}         
          & \emph{(Failed)} & 37.74 & \textbf{39.50} & 36.72 & 36.51 & 36.77 & 37.04 & 37.30\\ \midrule
        FedAVG &   \multicolumn{8}{c}{\textbf{36.30} (\emph{$\mu$=0.0})} \\ \midrule
        FedBABU &   \multicolumn{8}{c}{\textbf{45.73}} \\ \midrule
        FedFN &   \multicolumn{8}{c}{\textbf{46.98}} \\ \bottomrule
        
        \end{tabular}
        }
        \label{apptab:cifar100_fedfr}
\end{table}
\newpage
\subsection{FN in the Centralized Learning}
\begin{figure}[h!]
    \centering    
    \includegraphics[width=1.0\textwidth]{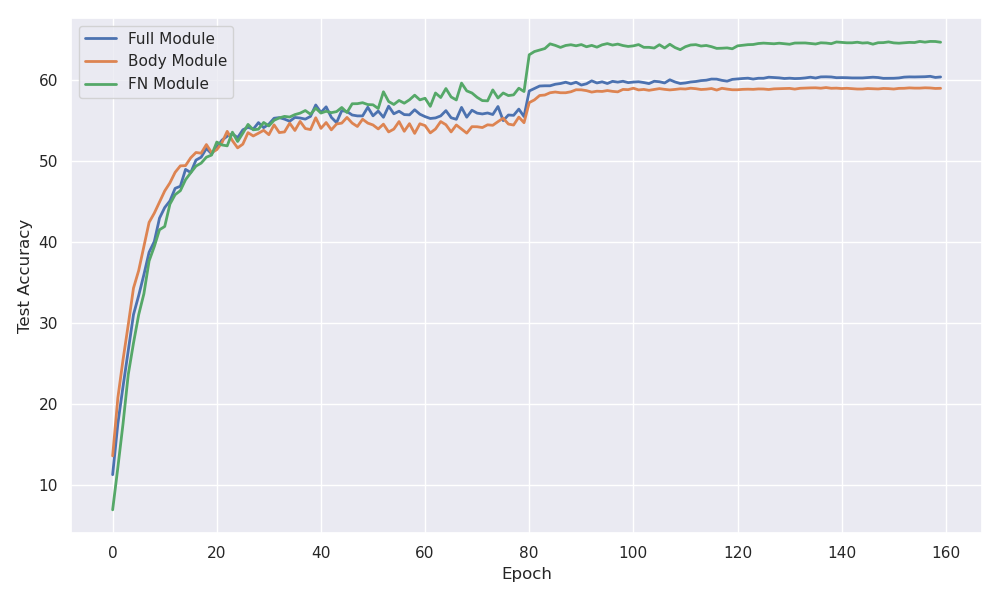}
    \caption{Module Comparison in Centralized Learning}
    \label{fig:centralized_acc_cifar100}
\end{figure}
In FedBABU~\cite{oh2021fedbabu}, the authors comprehensively evaluate the performance of the Full and Body modules in a centralized learning setup. Expanding upon their analysis, we incorporate the FN module to assess its performance in comparison to that of the Full and Body modules. To ensure fair comparisons, we replicate the experimental settings outlined in \cite{oh2021fedbabu}. The experiments are carried out on the CIFAR-100 dataset, utilizing the MobileNet architecture. The results of the conducted experiments are depicted in Figure~\ref{fig:centralized_acc_cifar100}.

As expected, our evaluation reveals a slight decline in performance within the Body module compared to the Full module, aligning with the findings reported in \cite{oh2021fedbabu}. This performance difference can be attributed to the partial update constraint imposed on the model during the training of the Body module. In contrast, the FN module, despite incorporating the constraint of feature normalization, exhibits significant improvements over the Full module. This enhancement is attributed to the capacity of FN module to empower the model to acquire more effective and discriminative representations, thereby enhancing overall performance.

\newpage

\end{document}